\pdfoutput=1

\documentclass[11pt]{article}

\usepackage[]{emnlp2023}

\usepackage{times}
\usepackage{latexsym}

\usepackage[T1]{fontenc}

\usepackage[utf8]{inputenc}

\usepackage{microtype}
\usepackage{times}
\usepackage{latexsym}
\usepackage{microtype}

\usepackage{inconsolata}

\usepackage{soul}
\soulregister\cite7
\soulregister\ref7 
\soulregister\citep7 
\soulregister\citet7 
\soulregister\pageref7 

\usepackage{makecell}
\usepackage{fontawesome}
\usepackage{makecell}

\usepackage{graphicx}
\usepackage{booktabs}
\usepackage{multirow}
\usepackage{multicol}
\usepackage{amssymb}
\usepackage{graphicx}
\usepackage{multirow}
\usepackage{subfig}
\usepackage{geometry} 
\usepackage{marvosym}

\title{Retentive or Forgetful? Diving into the Knowledge Memorizing Mechanism of Language Models}

\author{Boxi Cao${}^{1,3}$, Qiaoyu Tang${}^{1,3}$, Hongyu Lin${}^{1}$\textsuperscript{\Letter}, Shanshan Jiang${}^{4}$, Bin Dong${}^{4}$, \\ \textbf{Xianpei Han}${}^{1, 2}$\textsuperscript{\Letter}, \textbf{Jiawei Chen}${}^{1,3}$, \textbf{Tianshu Wang}${}^{1,3}$,\textbf{Le Sun}${}^{1,2}$\\
${}^{1}$Chinese Information Processing Laboratory ~ ${}^{2}$State Key Laboratory of Computer Science \\
Institute of Software, Chinese Academy of Sciences, Beijing, China\\
${}^{3}$University of Chinese Academy of Sciences, Beijing, China \\
${}^{4}$Ricoh Software Research Center (Beijing) Co., Ltd. \\
{\tt \{boxi2020,hongyu,xianpei\}@iscas.ac.cn} 
}

\begin{document}
\maketitle
\begin{abstract}
Memory is one of the most essential cognitive functions serving as a repository of world knowledge and episodes of activities. 
In recent years, large-scale pre-trained language models have shown remarkable memorizing ability.
On the contrary, vanilla neural networks without pre-training have been long observed suffering from the catastrophic forgetting problem. 
To investigate such a \emph{retentive-forgetful} contradiction and understand the memorizing dynamic mechanism of language models, we conduct thorough experiments by controlling the target knowledge types, the learning strategies and the learning schedules.
We find that: 1) \emph{Vanilla language models without pre-training are forgetful}; 2) \emph{Pre-training leads to retentive language models}; 3) \emph{Knowledge relevance and diversification significantly influence the memory formation}. 
These conclusions are useful for understanding the abilities of pre-trained language models and shed light on designing and evaluating new learning and inference algorithms of language models.
\end{abstract}

\section{Introduction}

Memory is one of the most essential cognitive functions serving as a repository of world knowledge and episodes of activities~\citep{squire1986mechanisms,duch2008cognitive}. 
Recent years have witnessed the boom of pre-trained language models (PLMs)~\citep{devlin2018bert,brown2020language,raffel2020exploring,touvron2023llama}, and the strong ability to memorize information in training data is considered a core factor for their success. 
For example, many previous works have shown that PLMs are retentive, which can effectively exploit training data and memorize information, including factual knowledge~\citep{petroniLanguageModelsKnowledge2019,jiangHowCanWe2020,cao-etal-2021-knowledgeable}, linguistic knowledge~\citep{linOpenSesameGetting2019,tenneyWhatYouLearn2019,ettingerWhatBERTNot2020}, commonsense knowledge~\citep{forbesNeuralLanguageRepresentations2019,Zhou2020EvaluatingCI} and narrative knowledge~\cite{carlini2021extracting,mccoy2021much,carlini2022quantifying}. 
Such strong memorizing ability is contributed to its over-parameterized neural network architectures, and has been regarded as one of fundamental reasons for their superiors performance across almost all NLP tasks~\citep{feldman2020does,DBLP:journals/corr/abs-2112-09153,tirumala2022memorization,tanzer-etal-2022-memorisation,cao2024life}. 

\begin{figure}[!t]
\centering
    \includegraphics[width=\columnwidth]{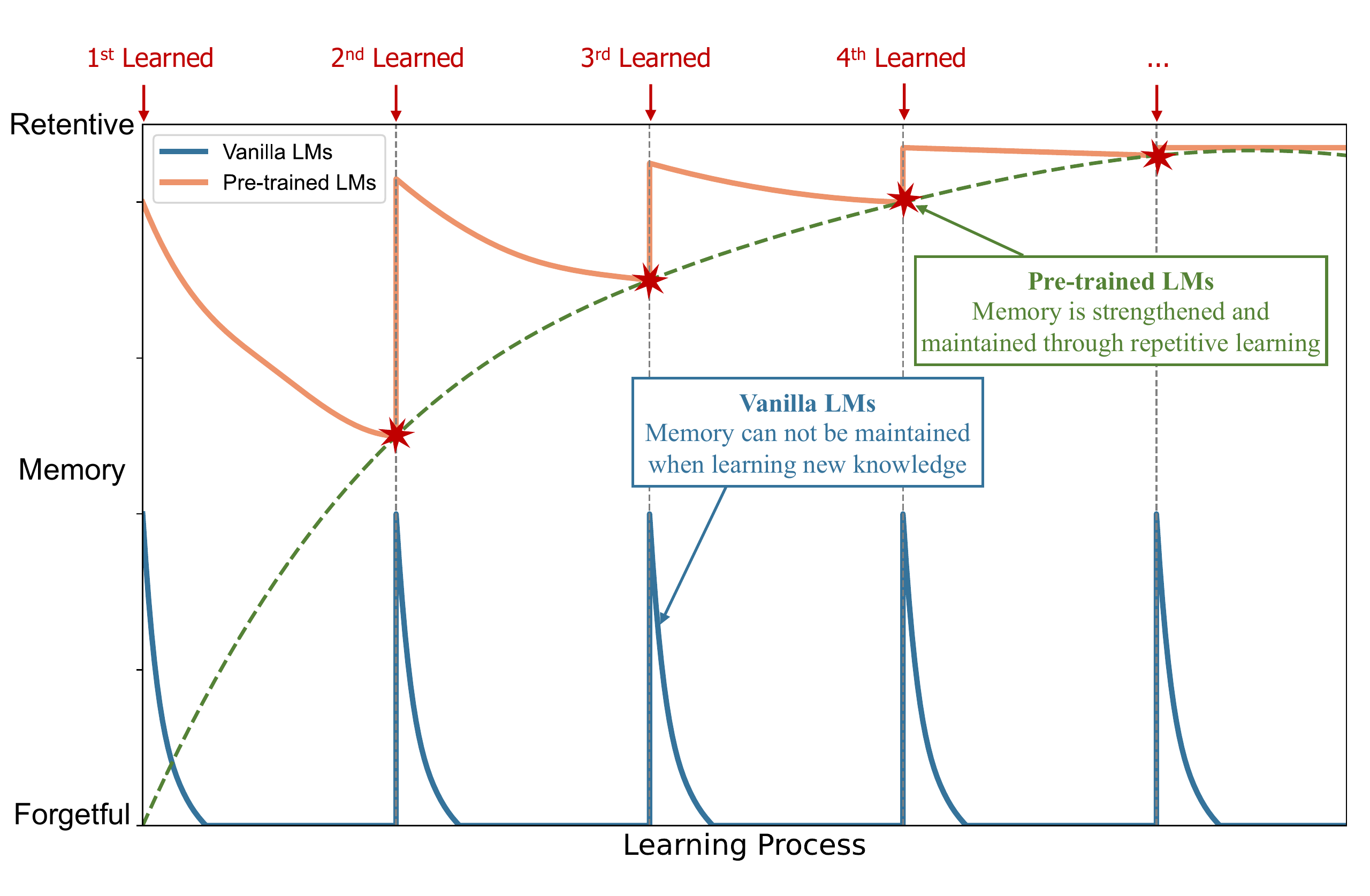}
    \caption{The memorizing dynamics of vanilla language models\protect\footnotemark and pre-trained language models. In the experiments, the factual knowledge is learned in a repetitive, group by group manner. Therefore, "1st learned" and "2st learned"indicates the time step where the target knowledge type is firstly or secondly learned respectively.}
    \label{fig:head}
\end{figure}

On the contrary, \textbf{\textit{catastrophic forgetting}} of neural networks has troubled the machine learning community for decades~\citep{mccloskey1989catastrophic,mcclelland1995there,french1999catastrophic}. That is, neural networks have long been observed with the tendency to abruptly and drastically forget previously learned knowledge upon learning new knowledge. 
\footnotetext{Vanilla language model refers to \textit{random initialized language model without pre-training} in this paper.}
Catastrophic forgetting is regarded as the most critical reason to prevent neural network architectures being successful models of memory~\citep{goodfellow2013empirical}. 
To overcome such forgetful nature, many techniques have been derived and investigated, such as continual learning~\citep{zenke2017continual,li2017learning, de2021continual}, life-long learning~\citep{richardson2008critical,chen2018lifelong,parisi2019continual} and multi-task learning~\citep{ruder2017overview,zhang2021survey}. 
Unfortunately, catastrophic forgetting has still been observed in many neural network learning processes across different kinds of tasks and architectures~\cite{pfulb2019comprehensive,ramasesh2021effect}.

Such a \textbf{\emph{retentive-forgetful}} observation raises an interesting contradiction about the memorizing mechanism of neural network-based language models. 
On the one hand, pre-trained language models do memorize the knowledge in a large amount of data, which brings to the success of large-scale PLMs. 
On the other hand, when facing many different downstream tasks, over-parameterized neural networks still show their nature of catastrophic forgetting, and are unable to persistently memorize knowledge from different sources. 
In this paper, we want to dive into such contradiction, and discover the underlying memorizing mechanisms that make the forgetful neural network architectures become retentive pre-trained language models. 
Specifically, this paper will explore the following three critical issues of the memorizing mechanism of language models:
\begin{itemize}
    \item Does the memory mechanism of language models follow specific patterns?
    \item How does pre-training affect the memorizing abilities of language models?
    \item What are the underlying factors that influence the memorization of language models?
\end{itemize}

To answer the above three questions, 
this paper uses the acquisition of factual knowledge as an empirical testbed to investigate the memorizing dynamics of current representative language model architectures. 
Specifically, we group the factual knowledge data according to different knowledge types, gradually and periodically train large-scale neural network-based language models to learn different types of factual knowledge, and then observe the models' memorizing dynamics of the previously learned knowledge at different time scales. 
For each type of knowledge serving as an experimental subject, we derive its \emph{forgetting curves}, which reflect the memorizing and forgetting dynamics of this kind of knowledge throughout the entire knowledge acquisition procedure. 
By summarizing evidence from forgetting curves of different models and knowledge types, we find that the memorizing dynamics of neural network-based LMs exhibit the following patterns which is illustrated in Figure~\ref{fig:head}, and are influenced by several critical factors:
 \begin{enumerate}
     \item \textbf{Vanilla language models are forgetful.} 
        We find vanilla language models (i.e., random initialized language models without pre-training) exhibit the catastrophic forgetting nature and can not persistently memorize the learned knowledge. 
        Specifically, forgetting curves of vanilla language models show that the knowledge are temporarily stored and with a clear periodicity.
        As new knowledge arrived, the previously learned knowledge is quickly forgotten, and repetitive learning hardly benefit the memory process.
        In other words, the memorizing dynamics of vanilla LMs exhibit characteristics of short memory duration and limited memory capacity.
    \begin{figure*}[!tp]
\centering
    \includegraphics[width=0.95\textwidth]{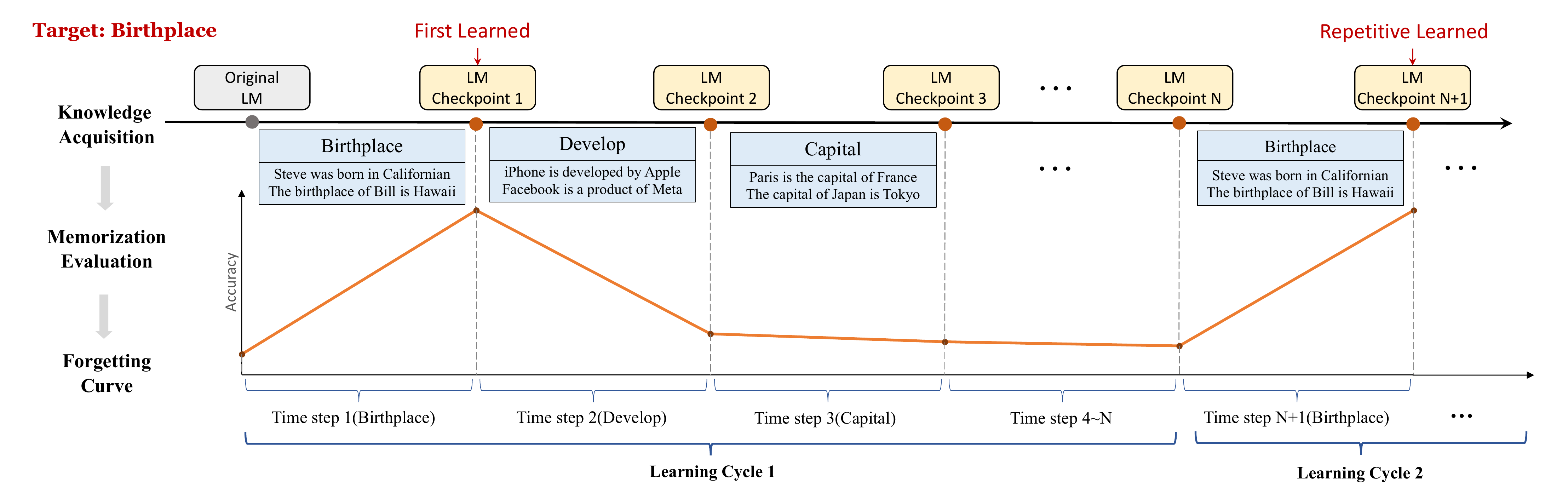}
    \caption{The illustrated factual knowledge acquisition testbed. During the knowledge acquisition process, each time step represents a group of knowledge being learned, and each learning cycle denotes a sequentially learning process of all knowledge groups in one epoch. Once a group of knowledge has been learned, we save a checkpoint and evaluate the current model performance on target knowledge type. Finally, the forgetting curve is plotted by summarizing the performance variation throughout the whole learning process which involves multiple learning cycles.}
    \label{fig:setup}
\end{figure*}
    \item \textbf{Pre-training leads to retentive language models.} 
        We find that pre-trained language models show dramatically different memorizing patterns from vanilla language models. Specifically, forgetting curves of PLMs show that the knowledge can be gradually memorized by the models throughout repetitive learning.
        Moreover, language models that have been pre-trained for a longer time have stronger memory ability, which reveals that pre-training is at the core of the forgetful to retentive transformation.
    \item \textbf{Knowledge relevance and diversification significantly influence memory formation.}
        We find that for pre-trained language models, there is a memory competition between highly related knowledge, i.e., when the newly learned knowledge is highly relevant to previously learned knowledge, it will undermine the memory of the previously learned knowledge retained in the model.
        Furthermore, in the early stage of memorizing process of PLMs, there frequently occurs \emph{memory collapse}, where the memory of previously learned knowledge sometimes seems to be rapidly disappeared but can be quickly recovered if a new, unrelated type of knowledge is further learned.
        We find that the occurrence of such memory collapse is highly related to the diversity of newly learned knowledge.
 \end{enumerate}

The remaining part of this paper is organized as follows. In Section~\ref{sec:setup}, we will briefly introduce our factual knowledge acquisition testbed and describe how to acquire the forgetting curves for each knowledge.
Section~\ref{sec:vanilla}-\ref{sec:corr} discuss our conclusions in detail respectively, including the corresponding supported findings and experiments.
At last, we will introduce the related works (\S \ref{sec:related}), as well as overall conclusions and discussions (\S \ref{sec:conc}).

\section{Factual Knowledge Acquisition Testbed}
\label{sec:setup}
To investigate the memorizing dynamics of language models, this paper leverages factual knowledge acquisition as a testbed, which aims to equip language models with real-world factual knowledge. 
A fact (e.g., <Steve Jobs, birthplace, California>) is an assertion containing a subject, an object, and the relation between them. 
To acquire factual knowledge, we convert facts into natural language expressions, and train language models using their corresponding \textit{pre-training objectives} (e.g., masked language modeling for BERT~\citep{devlin2018bert} and causal language modeling for GPT-2~\citep{radfordLanguageModelsAre2019}). To evaluate how well language models memorize a set of facts, we probe them using the corresponding factual knowledge queries (e.g., Steve Jobs was born in \_?) and check their answers. 

\paragraph{Dataset.} 
We construct our experiment testbed based on LAMA~\citep{petroniLanguageModelsKnowledge2019}, a well representative factual knowledge benchmark derived from Wikidata~\citep{vrandevcic2014wikidata}.
We select the 23 most common relations from LAMA, and regard each relation as a specific type of knowledge for LMs to learn. 
For each knowledge type, we randomly sample 10,000 facts from Wikidata, and generate 5 natural language sentences for each fact using 5 semantic-equivalent prompts~\citep{elazar2021measuring,cao-etal-2022-prompt}. 
For example, the sentences of <Steve Jobs, birthplace, California> may be 
``Steve Jobs was born in California'', ``the birthplace of Steve Jobs is California'', etc. 
In total, the dataset collects 23 groups of factual knowledge, with each group containing 50,000 sentences of a specific knowledge type, and each group of knowledge is sequentially sent to neural network-based language models for learning.


\paragraph{Language Model Architectures.}
We conduct experiments on transformer-based language model architectures, including BERT~\cite{devlin2018bert} and GPT-2~\cite{radfordLanguageModelsAre2019}. 
Due to page limitation, the results in the main body of paper are on BERT architecture without special declaration (BERT-base and BERT-large show the same results), and GPT-2 architecture reaches the same conclusions which can be refered in the appendix. 
We use Adam for optimization with learning rate of $1e-4$, $\beta_1=0.9, \beta_2=0.999$.

\paragraph{Knowledge Acquisition and Forgetting Curve.} 
Figure~\ref{fig:setup} illustrates the detailed knowledge acquisition process. 
Following relevant studies in psychology~\citep{wixted1991form,custers2010long,murre2015replication}, factual knowledge is learned in a repetitive, group-by-group manner by LMs in our experiments. 
Each group of data contains all derived instances of one knowledge type, once a group of knowledge has been learned, we save a checkpoint and 23 checkpoints will be saved in one epoch (i.e., learning cycle).

\begin{figure*}[!tp]
    \centering
    \includegraphics[width=0.9\textwidth]{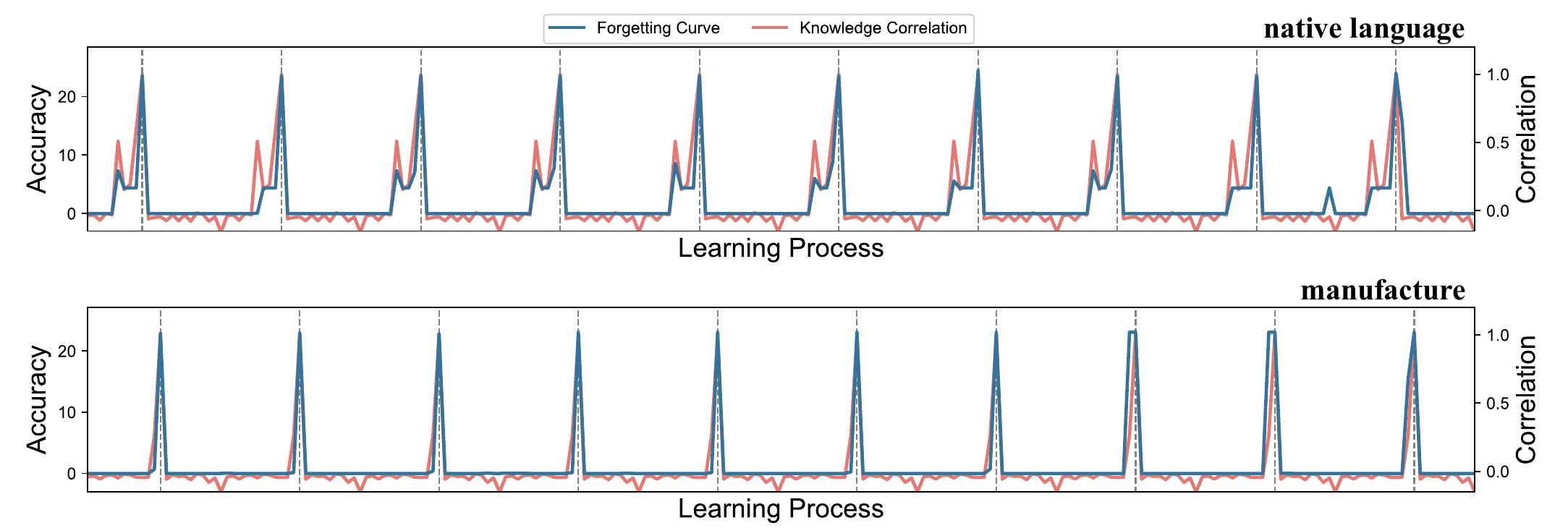}
    \caption{The forgetting curves and answer correlation curves of 
    knowledge \texttt{naive language} and \texttt{manufacture} on vanilla LMs. In the forgetting curve, each dashed line points to a time step when target knowledge is being repeatedly learned, with the middle portion representing a complete learning cycle. Each data point represents the knowledge probing performance on target knowledge type after a new knowledge type is learned. In the knowledge correlation curves, each data point denotes the Pearson correlation coefficient between the answer distributions of current step learned knowledge type and target knowledge type.}
    \label{fig:dynamic_curve}
\end{figure*}

\begin{figure}[!tp]
\centering
    \includegraphics[width=\columnwidth]{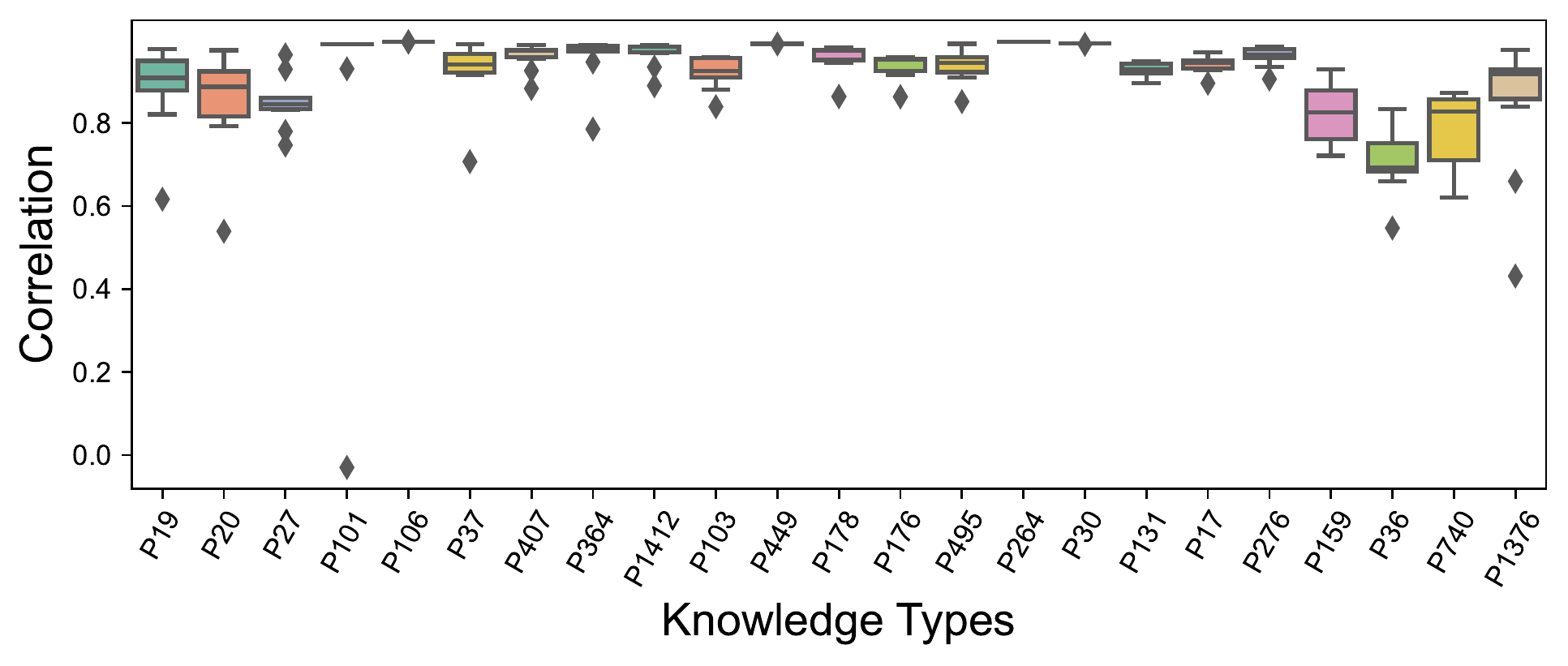}
     \caption{The Pearson correlation coefficients between the forgetting curve and answer correlation curve for each knowledge type. Each box represents the distributions of correlation coefficient on different learning cycles.}
    \label{fig:corr_box}
\end{figure}

To investigate the memorizing dynamics of LMs, we leverage \emph{forgetting curve}~\citep{ebbinghaus2013memory} as an effective toolkit.
To obtain the forgetting curve of a specific knowledge type: 
(1) we first evaluate its memorization on all checkpoints by probing with its corresponding factual knowledge queries (e.g., query BERT with \emph{Steve Jobs was born in [MASK]}); 
(2) the forgetting curve is plotted by summarizing its performance variation throughout the whole learning process.
Figure~\ref{fig:dynamic_curve} demonstrate several forgetting curves: 
the horizontal axis represents the learning process, where each time step represents a group of factual knowledge being learned, and each dashed line points to a time step when target knowledge is being repeatedly learned.
The vertical axis of the forgetting curve represents the performance of LM on the target knowledge type at the current time step, which reflects the memory of the LM for that type of knowledge.
In this way, the forgetting curve of a target knowledge type reflects its memorizing dynamics in a LM. 

\paragraph{Notations.} We refer to the knowledge type of a forgetting curve as its \emph{target knowledge type}, the sequentially learning process of all knowledge groups in one epoch as a \emph{learning cycle}, and the learning of a group of knowledge corresponds to a \emph{time step}, the group of knowledge learned at the most recent time step as \emph{newly learned knowledge}.

\section{Vanilla Language Models are Forgetful}
\label{sec:vanilla}

\textbf{Conclusion 1.} \emph{Vanilla language models are forgetful. Their memorizing dynamics exhibit a pattern similar to short-term memory.}

In this section, we apply factual knowledge acquisition testbed to vanilla language models, and analyze the forgetting curves of different knowledge types.
Throughout the experiments, we find that vanilla language models suffer severely from catastrophic forgetting: the models mostly only memorize the newly learned knowledge and forget the previously learned knowledge. 
As a result, the performance on a specific type of knowledge shows an obvious periodicity: language models achieve the best performance on the target knowledge type learning time step, and the performances at other time steps significantly degrade and are completely determined by the knowledge correlation between the target type and the latest learned type. 
Besides, repetitive learning cannot improve the memorization. 
Such short duration, limited capacity phenomena is very similar to the short-term memory mechanism in psychology, where the memory can only hold a small amount of information in active with a short interval~\citep{miller1956magical,atkinson1971control,goelet1986long} and are susceptible to be interfered by other memories. In the following, we will illustrate the experiment findings to reach the above conclusions.

\subsection{Limited Memory Duration}

\begin{figure*}[!ht]
    \centering
    \includegraphics[width=0.9\textwidth]{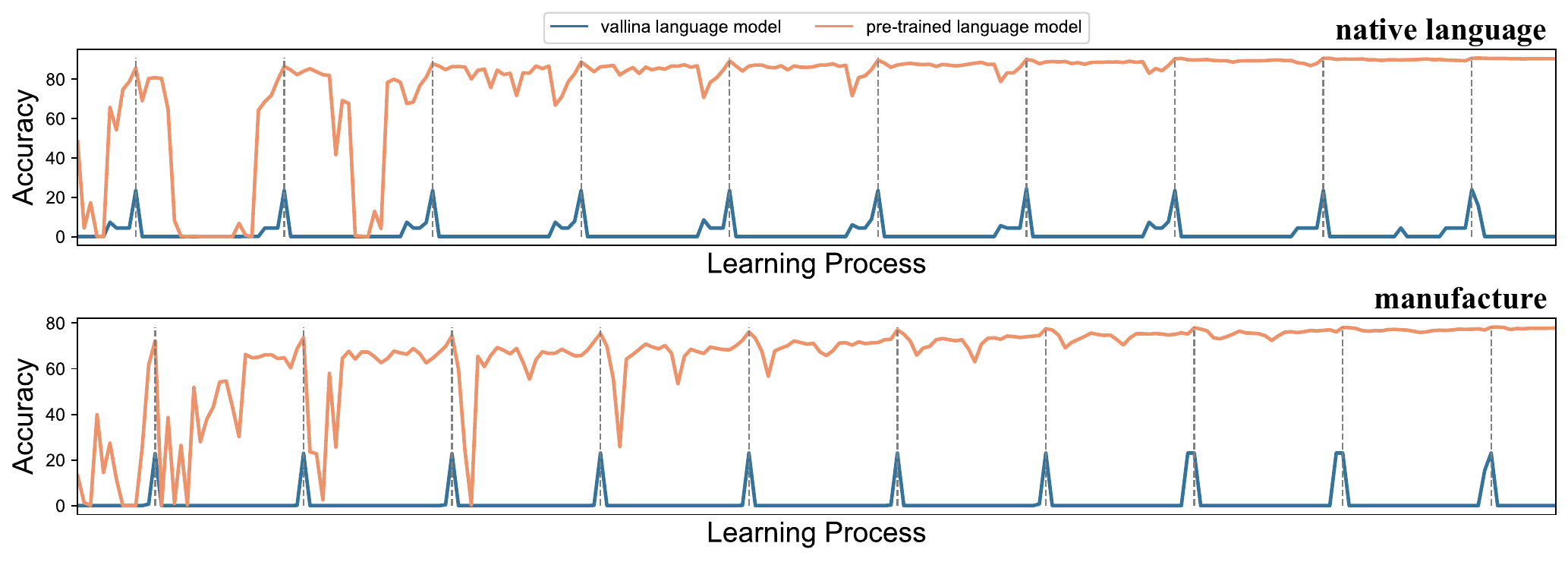}
    \caption{The forgetting curves of \texttt{native language} and \texttt{manufacture} on both vanilla and pre-trained LMs.}
    \label{fig:pretrain_curve}
\end{figure*}


\begin{figure*}[!ht]
\centering
    \includegraphics[width=0.9\textwidth]{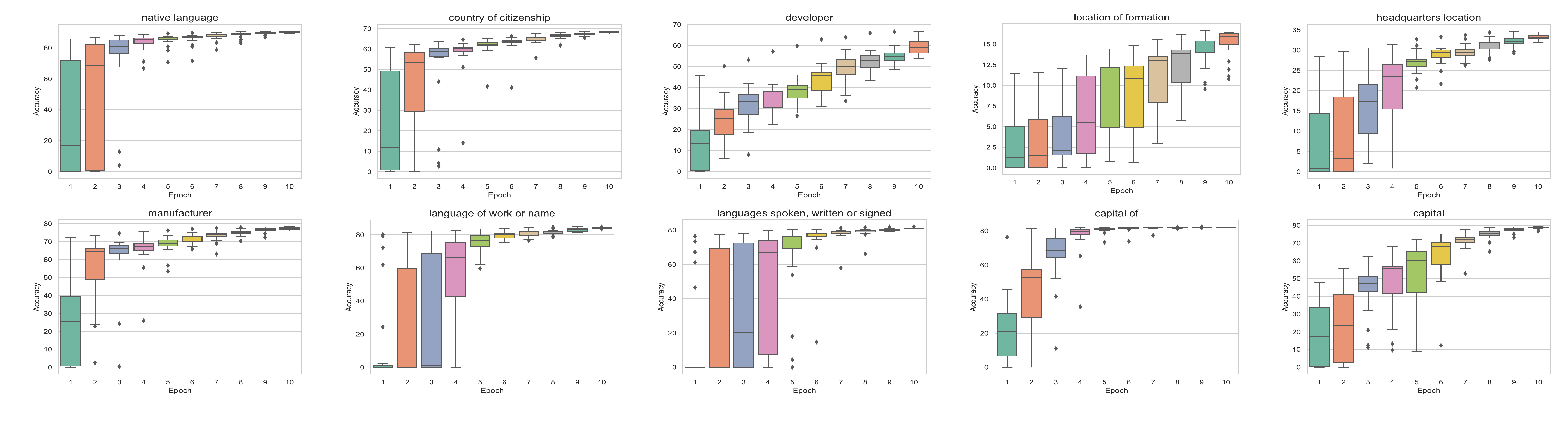}
    \caption{The box plot of performance change through learning cycles. Each box reflects the performance distribution of the target knowledge type in one learning cycle.}
    \label{fig:continue_box}
\end{figure*}

\textbf{Finding 1.} \emph{Vanilla language models will rapidly forget retained knowledge when new knowledge emerges, and repetitive learning has minimal impact on the effectiveness of knowledge retention.}

To show this, we investigate the dynamics of forgetting curves for vanilla language models. Figure~\ref{fig:dynamic_curve} demonstrates the forgetting curves of two representative knowledge types including \texttt{native language} and \texttt{manufacture}. 
More results and details of other knowledge types and language models are in the appendix, which demonstrate the same tendency. 

From Figure~\ref{fig:dynamic_curve}, we can see that the performance of vanilla language models shows an apparent periodicity: 
1) For the target knowledge type, the model achieves the best performance when the target knowledge type has just been learned, which indicates that the model can memorize the knowledge at its learning step; 
2) The performance quickly degrade even to zero once another kind of knowledge is learned, which confirms the existence of catastrophic forgetting; 
3) The entire learning and memorizing process is cyclical, and repetitive learning cannot effectively improve the memory duration of the target knowledge. 
These results indicate that there are no long-term memory effects in vanilla language models. In general, vanilla language models exhibit a short-term, forgetful memory pattern with very limited memory duration.


\subsection{Limited Memory Capacity}

\textbf{Finding 2.} \emph{The performance of vanilla language models is mostly determined by the knowledge learned in the nearest step, and the impact of knowledge learned from earlier steps is very limited.}

From Figure~\ref{fig:dynamic_curve}, we find the performance of target knowledge mostly quickly drops to zero once new knowledge are learned; this verifies the limited memory capacity of vanilla language models. 
However, we also observe some cases where language models can still retain a relative performance when new knowledge are learned. 
To investigate this, we estimate the knowledge correlations between the answer distribution of current step learned knowledge and the answer distribution of the target knowledge type. 
The results are illustrated in red lines in Figure ~\ref{fig:dynamic_curve}, which shows a clear and strong connection between the memorizing performance and the knowledge correlation: 1) The stronger the correlation between newly learned knowledge and the target knowledge, the better the performance on target knowledge; 2) the cycles of the correlations and that of the performance are completely consistent, highlighting the pivotal role of knowledge correlation in determining the performance of the target knowledge for vanilla language models.

To further quantify these findings, we compute the Pearson correlation coefficients of these two series (forgetting curve and knowledge correlation curve) on every target knowledge type. Figure~\ref{fig:corr_box} shows the box plots of coefficients. We can see that on almost all target knowledge types, the average correlation coefficient surpasses 0.9, displaying a consistently low standard variance across each learning cycle. 
All these results demonstrate that the newly learned knowledge, rather than the knowledge learned in earlier steps, dominates the performance, which indicates that the memory capacity of vanilla language models is quite limited.

\section{Pre-training Leads to Retentive Language Models}
\label{sec:pretrain}

\textbf{Conclusion 2.} \emph{Pre-training is at the core of the forgetful to retentive transformation. Consequently, pre-trained language models are retentive which exhibits a long-term memory pattern. }

To show this, we apply factual knowledge acquisition testbed to pre-trained language models such as BERT and GPT-2. 
We find that the memorizing pattern of pre-trained language models significantly diverges from that of vanilla language models: although there still exists forgetting at the beginning of learning, the memory of pre-trained language models will be significantly improved via repetitive learning. 
Furthermore, we find that a longer pre-training will result a stronger memorizing ability, i.e., target knowledge can be memorized faster and better.

The above findings reveal that pre-training is the key to transform forgetful vanilla language models with short memory duration and limited memory capacity, to the retentive pre-trained language models with long memory duration and large memory capacity. 
In the following, we will explain the above conclusions in detail.

\subsection{Pre-trained Language Models are Retentive}

\textbf{Finding 3.} \emph{Through repetitive learning, pre-trained language models can gradually remember and finally retain the learned knowledge.}

Figure~\ref{fig:pretrain_curve} shows two representative forgetting curves of both vanilla and pre-trained LMs on \texttt{native language} and \texttt{manufacture} knowledge, and more similar results on other knowledge types are shown in the appendix.
To make a fair comparison, we maintain identical model size, model architecture, learning schedule and hyper-parameters for both vanilla and pre-trained LMs in each experiments.

From Figure~\ref{fig:pretrain_curve}, we can see that pre-trained language models and vanilla language models follow totally different memorizing patterns. At the beginning stage of learning, the pre-trained LMs, like the vanilla LMs, will forget the knowledge learned in the earlier steps after learning new knowledge.
However, with the repetition of learning, PLMs can gradually remember all kinds of learned knowledge, and will not forget them once reaches a convergent state.
Besides, the memory capacity of PLMs is much larger than that of vanilla language model, and therefore PLMs can achieve much better performance on all kinds of knowledge.

To better quantify these findings, Figure~\ref{fig:continue_box} shows the box plots of the performance change across learning cycles, where each box plot contains 10 boxes corresponding to the first 10 learning cycles, and each box reflects the performance distribution of the target knowledge type in one learning cycle. From these box plots, we can see that the tendencies of knowledge memorization are very clear and consistent: 1) the average and the low-bound of performance in each learning cycle is gradually increased, which indicates that repetitive learning improves memorization; 
2) The performance variance in each learning cycle is gradually decreased, which indicates that the forgetting of knowledge is gradually weakening.
In summary, PLMs show significantly different memorizing patterns from vanilla language models, whose memory ability is with long duration and high capacity.

\begin{figure*}[!tp]
\centering
    \includegraphics[width=0.9\textwidth]{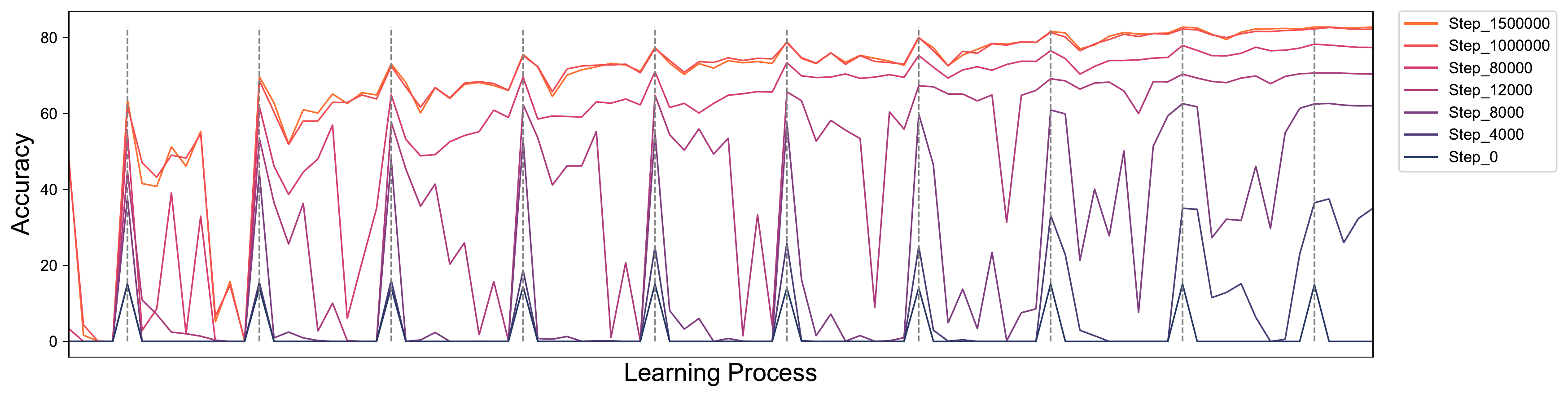}
    \caption{The forgetting curves on knowledge type \texttt{official language} of several pre-training language model checkpoints with different pre-training steps. Forgetting curves of other knowledge types will be shown in the appendix, which reach the same conclusions.}
    \label{fig:self_pretrain_curve}
\end{figure*}

\subsection{Pre-training Leads to Long-term Memorizing}
\label{subsec:pretrain}

\textbf{Finding 4.} \emph{Pre-training is the key to transforming short-term memory of vanilla language models into long-term memory of pre-trained language models.}

As shown above, the significant forgetful-retentive divergence between vanilla and pre-trained language models reveals the decisive role of pre-training in the formation of long-term memory ability.
To further understand how the pre-training procedure shapes the memory ability of language models, we investigate the memory dynamics of language models in different stages of pre-training.
Specifically, we pre-train a new BERT-based language model from scratch, collect language models at different pre-training checkpoints with sufficient step intervals, and then compare their memorizing dynamics over the same target knowledge type. 
We perform training on 2 Nvidia A100-80 GPUs for about 10 days, and follow the original pre-training process of BERT: using Wikipedia and Bookcorpus as datasets, training with batch size of 512 sequences for 1,500,000 steps, using Adam with learning rate of $1e-4$, $\beta_1=0.9, \beta_2=0.999$.

Figure~\ref{fig:self_pretrain_curve} shows the forgetting curve of knowledge type \texttt{citizenship} with different pre-training steps, and more knowledge types are presented in the appendix.
From this figure, we can clearly observe the formation process of memory ability along with the pre-training process. 
Firstly, the ability for long-term memory is formed in the early stage of pre-training: through 80,000 steps of pre-training, the memorizing pattern of LM shifts from forgetful short-term memory to retentive long-term memory. 
Secondly, the capacity of memory is affected by the depth of pre-training. That is, the longer the pre-training step, the larger the PLMs' memory capacity and the better the final memorizing performance. The above findings clearly show the critical role of pre-training for memorizing ability formation of NN-based language models. 


\begin{figure}[!tp]
\centering
    
    \includegraphics[width=\columnwidth]{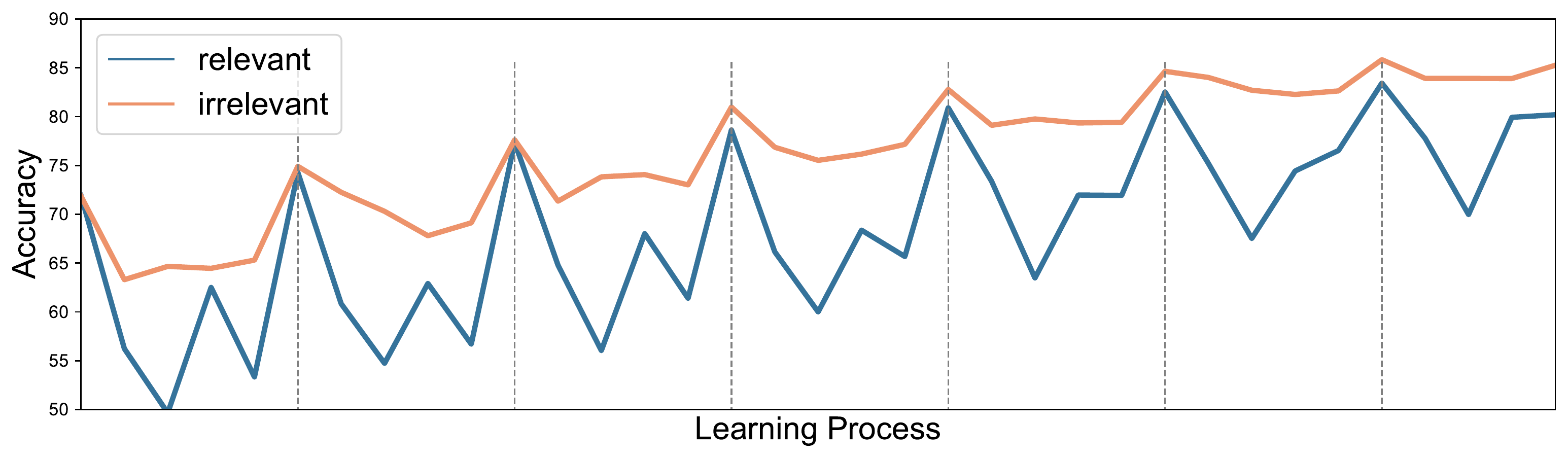}
    \caption{The forgetting curves of knowledge type \texttt{official language} learned with relevant knowledge and with irrelevant knowledge.}
    \label{fig:corr_probing_curve}
\end{figure}

\section{Knowledge Relevance and Diversification Affect Memory Formation}
\label{sec:corr}
\textbf{Conclusion 3.} \emph{The correlation between newly learned knowledge and retained knowledge, as well as the diversification of newly learned knowledge, will interfere with the memorization in pre-trained language models.}

From Figure~\ref{fig:pretrain_curve} and ~\ref{fig:self_pretrain_curve}, we can see that although PLMs exhibit a long-term, retentive memory pattern, the memory formation of a specific type of knowledge shows obvious fluctuations before final convergence. Specifically, the memorization performance of a knowledge type may dramatically diverge in different steps of the same learning cycle. More strikingly, we find that there frequently occurs \emph{knowledge memory collapse}, i.e., the memory of previously learned knowledge sometimes seems to be rapidly collapsed but can be quickly recovered in the next step. 

To deeply understand the causes of performance fluctuation and memory collapse, we explore the potential factors affecting the memory formation process of PLMs. 
Inspired by interference theory of forgetting~\citep{postman1961present,ceraso1967interference}, 
we intensively study the impact of knowledge correlations between the retained and the newly learned knowledge, as well as the characteristics of newly learned knowledge that may cause knowledge memory collapse. 
We find that there is a memory competition between highly related knowledge, and the emergence of memory collapse is highly related to the diversification of newly learned knowledge.

\subsection{Correlations brings Competitions}


\textbf{Finding 5.} \emph{There is a competition between the memorization of highly relevant knowledge types in pre-trained language models.}

To understand the high memorization performance fluctuation in Figure~\ref{fig:pretrain_curve} and ~\ref{fig:self_pretrain_curve}, we investigate the relation between the newly learned knowledge and the target knowledge, and find that 
the memory of a knowledge type will be undermined due to the competition of highly relevant knowledge types.

\begin{figure}[!tp]
\centering
    \includegraphics[width=0.9\columnwidth]{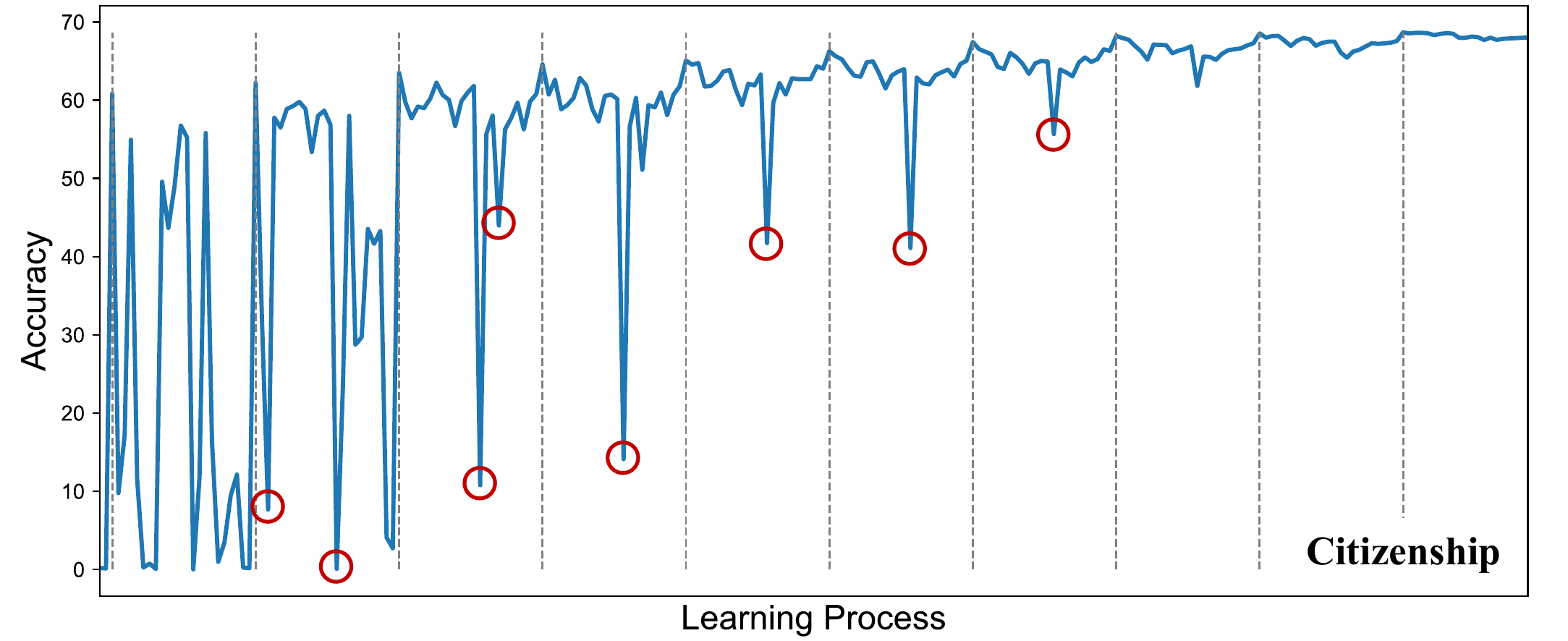}
    \caption{The memory collapse phenomenon for \texttt{citizenship}, which indicates the memory of target knowledge rapidly collapse but can quickly recovered in the next step. The memory singularities after warming up are marked in red.}
    \label{fig:singular}
\end{figure}

To verify these observations, we compare the forgetting curves correspondingly learned with relevant knowledge and with irrelevant knowledge. Figure~\ref{fig:corr_probing_curve} shows the forgetting curves of \texttt{official language} learned with a group of relevant knowledge types (e.g., \texttt{work language}, \texttt{native language}) and with a group of irrelevant knowledge types (e.g., \texttt{occupation}, \texttt{record label}). We can see that, the performance learned with relevant knowledge decreases more significantly than with irrelevant knowledge. That is, the newly learned correlated knowledge would interfere with the original memory in PLMs.

\subsection{Memorizing Singularity and its Causes}
\label{ssec:collapse}

\textbf{Finding 6.} \emph{The lack of knowledge diversity is a critical factor to the memory collapse in the early stage of learning process.}

We observe a striking \emph{memory collapse} phenomenon as shown in Figure~\ref{fig:singular}, i.e., at some singularities (red circles), memory of target knowledge type rapidly collapses but will be quickly recovered in the next step, and the memory collapse will gradually diminish along with repetitive learning.

To analyze the causes of memory collapse, we collect all singularities of memory collapse, and identify the knowledge type learned at the singularity step (more details are in the appendix).
We find that most of the memory collapses are caused by the knowledge with a very low answer diversity, e.g., \texttt{Continent} knowledge with only 7 possible answers. 
This observation enlightens us on whether the memorizing singularities are caused by the lack of diversity of newly learned knowledge.

\begin{figure}[!tp]
\centering
    \includegraphics[width=0.48\textwidth]{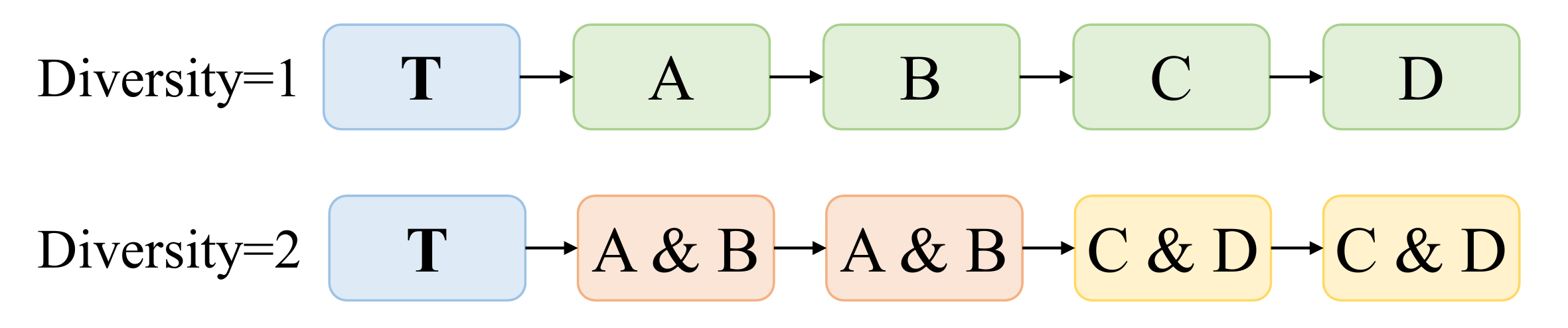}
    \caption{The knowledge memorization with different diversity settings, here T denotes target knowledge type, A/B/C/D denotes other knowledge types, and each box corresponds to a time step. We increase the diversity by sequentially mixing and randomly shuffling $K$ types of knowledge in one time step (except time step for T).}
    \label{fig:div}
\end{figure}

\begin{figure}[!tp]
\centering
    \includegraphics[width=\columnwidth]{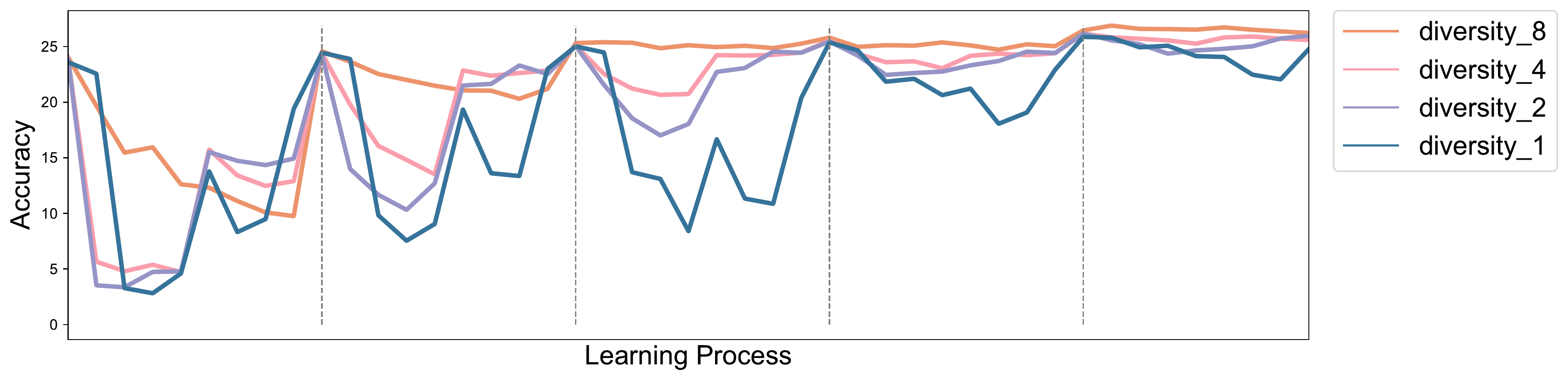}
    \caption{The forgetting curves of knowledge type \texttt{birthplace} with different knowledge diversities.}
    \label{fig:div_curve}
\end{figure}

To verify this hypothesis, we compare the forgetting curves using different diversity settings as shown in Figure~\ref{fig:div}.
Specifically, given a specific target knowledge type and the knowledge diversity degree $K$, we classify the rest of the data into groups of $K$ knowledge types, and then repetitively learns the target knowledge and these groups of knowledge. 
The data in the same group are randomly shuffled to ensure that  instances of different knowledge types can be included in each batch of learning.
In this way, each learning step will contains knowledge sampled from $K$ kinds of knowledge types, and therefore the diversity at each learning step can be roughly approximated as proportional to $K$.
Figure~\ref{fig:div_curve} shows the results of \texttt{Birthplace} with diversity degrees 1, 2, 4 and 8. 
We can see that the more diverse the newly learned knowledge, the faster the memorizing process converges. Furthermore, when diversity increases, the model collapse becomes shallower and appears less frequently. This confirms our hypothesis that knowledge diversification is critical to avoid memory collapse, and therefore plays a significant role in the memory formation process of PLMs.

\section{Related Work}
\label{sec:related}


Memory is one of the most essential cognitive functions~\citep{squire1986mechanisms,duch2008cognitive,eysenck2012attention}, and extensive studies have been conducted on both psychology and AI for decades~\citep{miller1956magical,gathercole1998development,mccloskey1989catastrophic}.
For neural networks, lots of studies have focused on the catastrophic forgetting problem~\citep{mcclelland1995there,french1999catastrophic,parisi2019continual}, which try to understand the reasons~\citep{french1991using,goodfellow2013empirical} and alleviate the forgetting~\citep{robins1995catastrophic,gepperth2016bio,zenke2017continual,li2017learning}.

Recent years have witnessed the remarkable memorizing ability of PLMs, and a lot of studies have conducted to PLMs' memorization, such as knowledge memory probing~\citep{petroniLanguageModelsKnowledge2019,jiangHowCanWe2020,cao-etal-2021-knowledgeable}, memorization ability quantification~\citep{mccoy2021much,carlini2022quantifying,jang2022towards}, memorization during training~\citep{tirumala2022memorization,tanzer-etal-2022-memorisation}, and the impact of pre-training on life-long learning~\citep{DBLP:journals/corr/abs-2112-09153}.
Several popular testbeds have been proposed, including factual knowledge probing~\citep{petroniLanguageModelsKnowledge2019,jiangXFACTRMultilingualFactual2020,kassner2021multilingual} or textual information memorization~\citep{carlini2021extracting,mccoy2021much,carlini2022quantifying}.
In this paper, we focus on the forgetful-retentive contradiction between vanilla and pre-trained LMs using factual knowledge acquisition as testbed, specifically, we investigate the knowledge memorizing dynamics and regularies of LMs, as well as the significant factors which influence memory formation.

\section{Conclusions and Discussions}
\label{sec:conc}
This paper investigates the memorizing mechanisms for both vanilla language models and pre-trained language models, and finds that: 1) Vanilla language models are forgetful; 2) Pre-training leads to retentive language models; 3) Knowledge relevance and diversification significantly influence memory formation.

These findings can benefit many other researches.
Firstly, the memorizing mechanisms of language models can provide a useful explanation to many previous studies such as the reason of catastrophic forgetting and the effect of pre-training. Secondly, by revealing several critical factors that affect PLMs' memorization, more stable and reliable learning algorithms can be designed for PLMs.
Finally, we find the memorization patterns of language models may be similar to human brains, which raises the opportunities to bridge research on AI and psychology.

This paper also raises many open problems for future study:
1) The underlying reasons why pre-training could lead to such a significant forgetful-retentive transition, are there any other synchronized transitions during pre-training?
2) Besides repetitive learning, pre-training and knowledge diversity, are there any other critical factors that affect memorizing ability?
3) What are the key differences between the brain's and language models' memorization? What abilities are required to achieve brain-like memorization?

\section*{Limitations}

This paper focuses on the factual knowledge acquisition, which is one of most representative memorization testbeds in both AI~\citep{petroniLanguageModelsKnowledge2019,jiangHowCanWe2020,jiangXFACTRMultilingualFactual2020,kassner2021multilingual} and psychology~\citep{tulving1998episodic,squire1998episodic}. In future work other memorization tasks such as text memorization can be further conducted. 

Due to the huge cost and the limitations of computing and storage resources (230 checkpoints required to be saved in each experiments), we have not yet conduct experiments on extremely large language models such as GPT-3~\citep{brown2020language} and LLaMA~\citep{touvron2023llama}. Meanwhile, due to the larger capacity of larger-scale models, we have reason to believe that our findings will remain consistent on these models as well, as they possess stronger memory capabilities.

\section*{Acknowledgements}
This work is supported by the Natural Science Foundation of China (No. 62122077 and 62106251) and the CAS Project for Young Scientists in Basic Research under Grant No.YSBR040.

\bibliography{new_simple}
\bibliographystyle{acl_natbib}

\appendix

\newpage

\section{Details of Knowledge Types}

\begin{table}[!tp]
\centering
\resizebox{0.48\textwidth}{!}{
\begin{tabular}{lrll}
\toprule
\textbf{Type} & \textbf{Relation ID}                         & \textbf{Relation Label}                 & \textbf{Example Prompt}                       \\ \hline
\multirow{5}{*}{Personal Information}  &P19      & place of birth                    & {[}X{]} was born in {[}Y{]}                   \\
                                       &P20      & place of death                    & {[}X{]} died in {[}Y{]}                       \\
                                       &P27      & country of citizenship            & The citizenship of {[}X{]} is {[}Y{]}         \\
                                       &P101      & field of work                     & {[}X{]} works in the field of {[}Y{]}         \\
                                       &P106      & occupation                        & {[}X{]} works as {[}Y{]}                      \\ \hline
\multirow{5}{*}{Languages}             &P37      & official language                 & The official language of {[}X{]} is {[}Y{]}   \\
                                       &P407      & language of work                  & {[}X{]} was written in {[}Y{]}                \\
                                       &P364      & original language of film         & The original language of {[}X{]} is {[}Y{]}   \\
                                       &P1412      & languages spoken                  & {[}X{]} used to communicate in {[}Y{]}        \\
                                       &P103      & native language                   & The native language of {[}X{]} is {[}Y{]}     \\ \hline
\multirow{5}{*}{Manufacture}           &P449      & original broadcaster              & {[}X{]} was originally aired on {[}Y{]}       \\
                                       &P178      & developer                         & {[}X{]} is developed by {[}Y{]}               \\
                                       &P176      & manufacturer                      & {[}X{]} is produced by {[}Y{]}                \\
                                       &P495      & country of origin                 & {[}X{]} was created in {[}Y{]}                \\
                                       &P264      & record label                      & {[}X{]} is represented by music label {[}Y{]} \\ \hline
\multirow{8}{*}{Location}              &P30      & continent                         & {[}X{]} is located in the continent {[}Y{]}   \\
                                       &P131      & located in the territorial entity & {[}X{]} is in the region of {[}Y{]}           \\
                                       &P17      & country                           & {[}X{]} is located in the country of {[}Y{]}  \\
                                       &P276      & location                          & {[}X{]} is located in {[}Y{]}                 \\
                                       &P159      & headquarters location             & The headquarter of {[}X{]} is in {[}Y{]}      \\
                                       &P36      & capital                           & The capital of {[}X{]} is {[}Y{]}             \\
                                       &P740      & location of formation             & {[}X{]} was founded in {[}Y{]}                \\
                                       &P1376      & capital of                        & {[}X{]} is the capital of {[}Y{]}    \\ \bottomrule         
\end{tabular}}
\caption{The details about the training relations.}
\label{tab:rels}
\end{table}

The training dataset in our experiments contain 23 relations from LAMA~\citep{petroniLanguageModelsKnowledge2019}, and each relation corresponds to 5 different semantically equivalent prompts which are collected from~\citet{elazar2021measuring,cao-etal-2022-prompt}.
The relations can be classified into four types including: personal information, language related, manufacture related and location information.
Table~\ref{tab:rels} shows the details for each relation.


\section{Forgetting Curves of Other Knowledge Types on BERT and GPT}

Figure~\ref{fig:more_forgetcurve} demonstrates the forgetting curves of more target knowledge types on both vanilla language models and pre-trained language models (including BERT-base and BERT-large), and Figure~\ref{fig:gpt_curve} shows the forgetting curves of vanilla and pre-trained GPT-2, which also reach the same conclusions in Section~\ref{sec:vanilla} and~\ref{sec:pretrain}.
Please kindly note that all forgetting curves show the same trends but a few knowledge types (country, location, continent) need more epochs to converge, therefore their forgetting curves don't reach a convergence in the presented first 10 learning epochs, but they all will reach the same conclusions after more learning epochs.

Furthermore, Figure~\ref{fig:more_pretraining_curve} demonstrate more comparison between the language model checkpoints during pre-training with different pre-training steps, which also confirm the conclusion of \emph{``pre-training is the key to transform short-term memory of vanilla language models into long-term memory of pre-trained language models''} in Section~\ref{sec:pretrain}.

\begin{table}[!tp]
\resizebox{\columnwidth}{!}{
\begin{tabular}{ll}
\toprule
\textbf{Relevant Knowledge}          & \textbf{Irrelevant Knowledge} \\
language of work or name             & occupation                    \\
original language of film or TV show & original broadcaster          \\
languages spoken, written or signed  & manufacturer                  \\
native language                      & record label                  \\ \bottomrule
\end{tabular}}
\caption{The relevant knowledge types and irrelevant knowledge types for target knowledge type \texttt{official language}.}
\label{tab:related_rels}
\end{table}

\section{Knowledge Relevance Analysis}

As we mentioned in Section~\ref{sec:corr}, we compare the forgetting curves correspondingly learned with a group of relevant knowledge and with a group of irrelevant knowledge. Table~\ref{tab:related_rels} shows the details for each group of knowledge, where we can clearly see the difference.
The group of relevant knowledge including knowledge types such as \texttt{language of work or name} and \texttt{native language}, which shares the similar relation semantics and answer distribution with the target knowledge type \texttt{official language}.
In comparison, the group of irrelevant knowledge including knowledge types such as \texttt{occupation} and \texttt{manufacturer}, which has no overlap in answer distribution with the target knowledge type \texttt{official language}.

\section{Memory Collapse Analysis}

\begin{table*}[!ht]
\centering
\resizebox{0.9\textwidth}{!}{
\begin{tabular}{lccc}
\toprule
\textbf{\makecell[l]{ Learned Knowledge \\ at Singularity }}       & \textbf{ \makecell[c]{Cause Collapse \\ Times}} & \textbf{\makecell[c]{Number of \\ Distinct Answers}} & \textbf{\makecell[c]{Entropy of \\ Answer Distribution}} \\ \hline
\textbf{continent}                               & \textbf{39}                   & \textbf{7}                        & \textbf{1.94}                        \\
country                                          & 26                            & 139                      & 5.45                                    \\
country of original                              & 4                             & 130                      & 5.09                                    \\
occupation                                       & 3                             & 210                      & 4.7                                     \\
official language                                & 2                             & 92                       & 4.24                                    \\
native language                                  & 2                             & 73                       & 3.81                                    \\
language of work or name                         & 2                             & 75                       & 2.39                                    \\
original network                                 & 2                             & 92                       & 4.04                                    \\
located in the territorial entity & 1                             & 1066                     & 8.22                                    \\
place of birth                                   & 0                             & 960                      & 8.37                                    \\
place of death                                   & 0                             & 793                      & 7.58                                    \\
country of citizenship                           & 0                             & 139                      & 5.24                                    \\
field of work                                    & 0                             & 800                      & 7.12                                    \\
original language of film or TV show             & 0                             & 75                       & 3.19                                    \\
languages spoken, written or signed              & 0                             & 75                       & 3.82                                    \\
developer                                        & 0                             & 156                      & 4.68                                    \\
manufacturer                                     & 0                             & 185                      & 5.02                                    \\
record label                                     & 0                             & 206                      & 4.45                                    \\
location                                         & 0                             & 1101                     & 7.68                                    \\
headquarters location                            & 0                             & 719                      & 7.32                                    \\
capital                                          & 0                             & 1078                     & 9.47                                    \\
location of formation                            & 0                             & 650                      & 7.74                                    \\
capital of                                       & 0                             & 561                      & 8.95                                    \\ \hline
Average                                          & 3.52                          & 407.91                   & 5.67                                   \\ \bottomrule
\end{tabular}
}
\caption{The details of learned knowledge types that cause memory collapse.}
\label{tab:collapse}
\end{table*}

As we mentioned in Section~\ref{ssec:collapse}, we observe a memory collapse phenomenon, i.e., at some singularities the memory of target knowledge type rapidly collapses but will be quickly recovered in the next step, and the memory collapse will gradually diminish along with repetitive learning.
To investigate the cause of such memory collapse, we collect all singularities of memory collapse, and identify the knowledge type learned at the singularity step.
Table~\ref{tab:collapse} shows the results of each learned knowledge type at singularity and the times it cause memory collapse.
And we can clearly see that the most memory collapses are caused by \texttt{Continent}. 
Compared \texttt{Continent} with other knowledge types, the most significant characteristic is that the answer diversity of \texttt{Continent}, as well as the entropy of its answer distribution, is substantially lower than other knowledge types. 
Specifically, for each distinct answer $e_i$ with frequency $f_i$ for relation $r$, the entropy $H_r$ of answer entity distribution for knowledge type $r$ can be calculated with:
\begin{equation}
    P_i = \frac{f_i}{\Sigma_{i} {f_i}}, H_r = -\Sigma_{i} P_i log_2 P_i.
\end{equation}
Table~\ref{tab:collapse} demonstrates the number of distinct answers and the entropy of answer distribution for each knowledge type.
We can clearly see the difference between knowledge type \texttt{Continent} and other knowledge types: 1) the answer set of \texttt{Continent} contains only 7 possible answers while other knowledge types contains at least 70, where the average across all knowledge types is 407.91. 2) the entropy of answer distribution for \texttt{Continent} is only 1.94 while the average entropy across all knowledge types is 5.67.
Such results are also understandable since there are only seven continents in the world.
To further verify our hypothesis, we calculate the prediction distribution of pre-trained language model every time when it is learned with the knowledge type \texttt{Continent}.
And we find that, at the memory singularities caused by \texttt{Continent}, the entropy for prediction distribution of PLMs would also significantly reduce, and PLMs tend to predict the continent as answer no matter what you ask. The prediction distribute would back to normal when the next relation is learned.
Using knowledge \texttt{Citizenship} as example, when PLM is learned with \texttt{Continent} and then query PLM with ``The citizenship of Isaac Newton is [MASK]'', the answer would be ``Europe'' instead of any country.



\begin{figure*}[!ht]
    \centering
    \subfloat[The forgetting curve of \texttt{country of citizenship}.]{\includegraphics[width=0.95\textwidth]{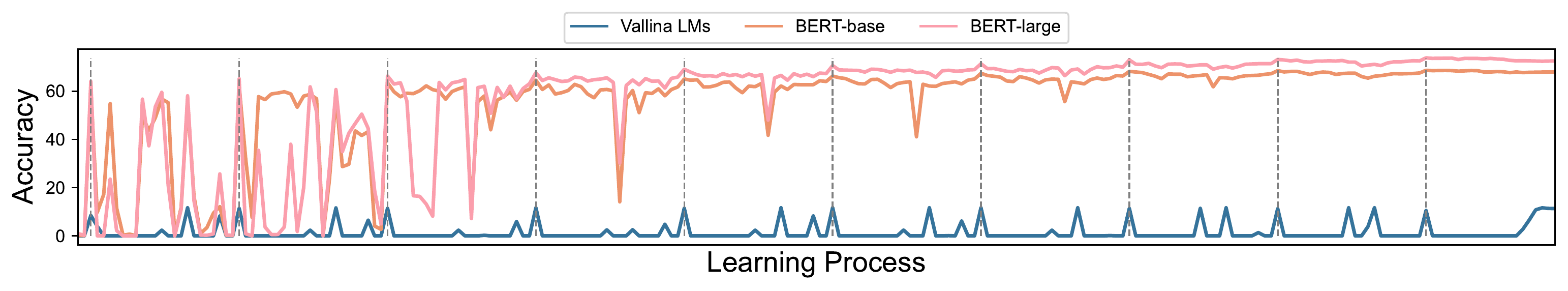}} \\
    \subfloat[The forgetting curve of \texttt{field of work}.]{\includegraphics[width=0.95\textwidth]{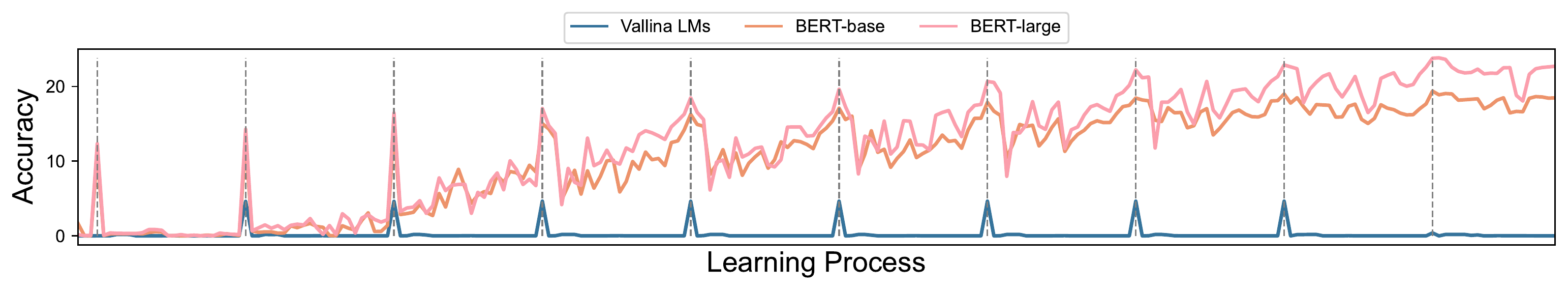}} \\
    \subfloat[The forgetting curve of \texttt{occupation}.]{\includegraphics[width=0.95\textwidth]{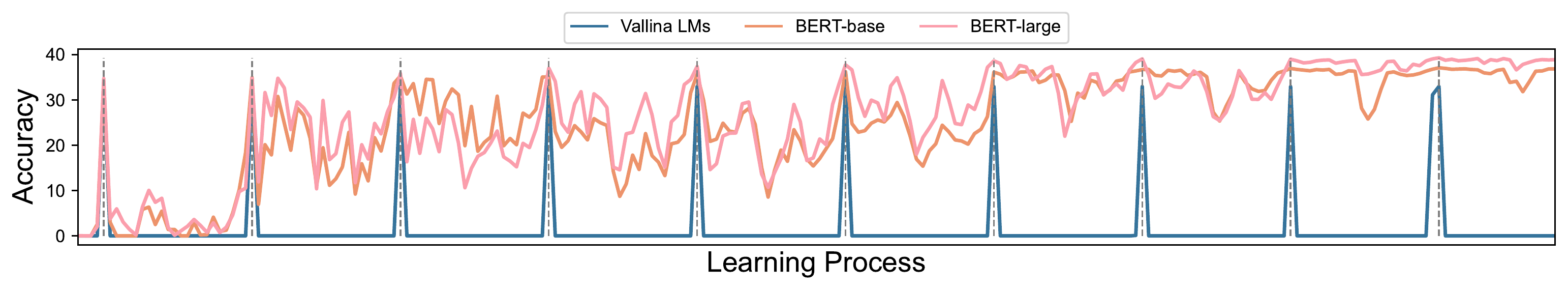}} \\
    \subfloat[The forgetting curve of \texttt{official language}.]{\includegraphics[width=0.95\textwidth]{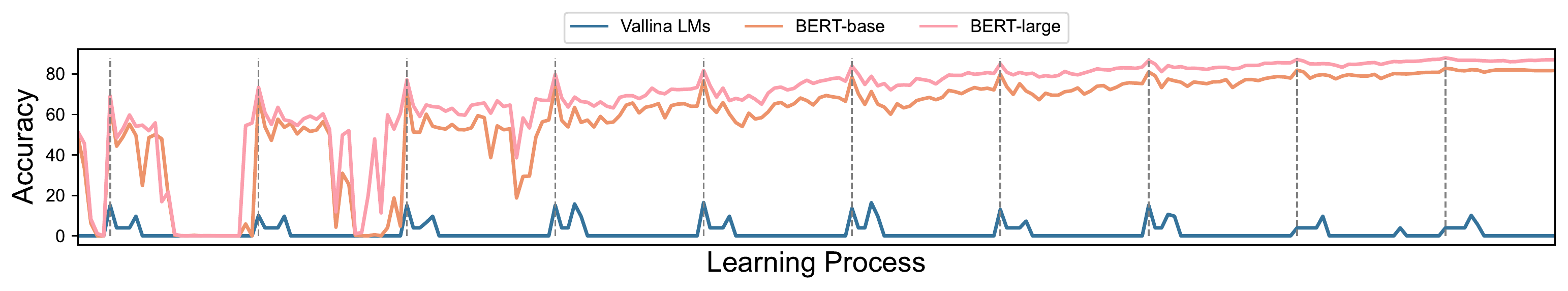}} \\
    \subfloat[The forgetting curve of \texttt{place of birth}.]{\includegraphics[width=0.95\textwidth]{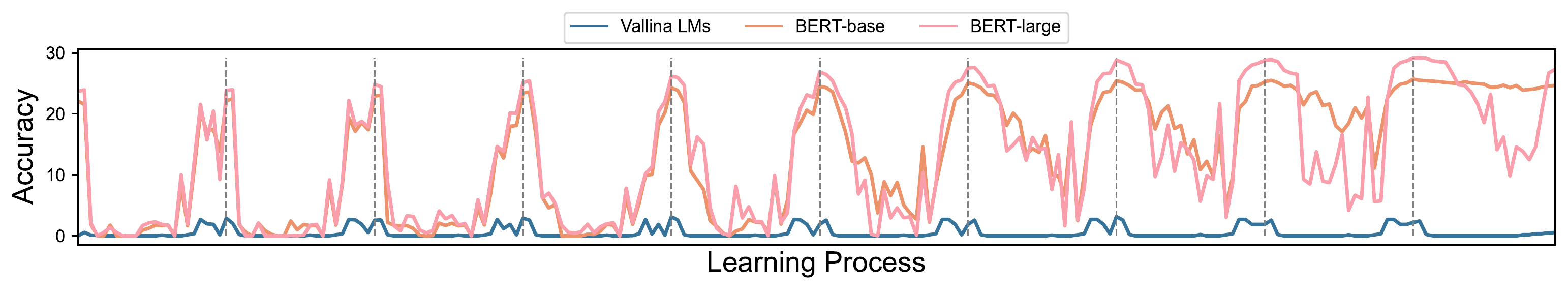}} \\
    \subfloat[The forgetting curve of \texttt{language of work}.]{\includegraphics[width=0.95\textwidth]{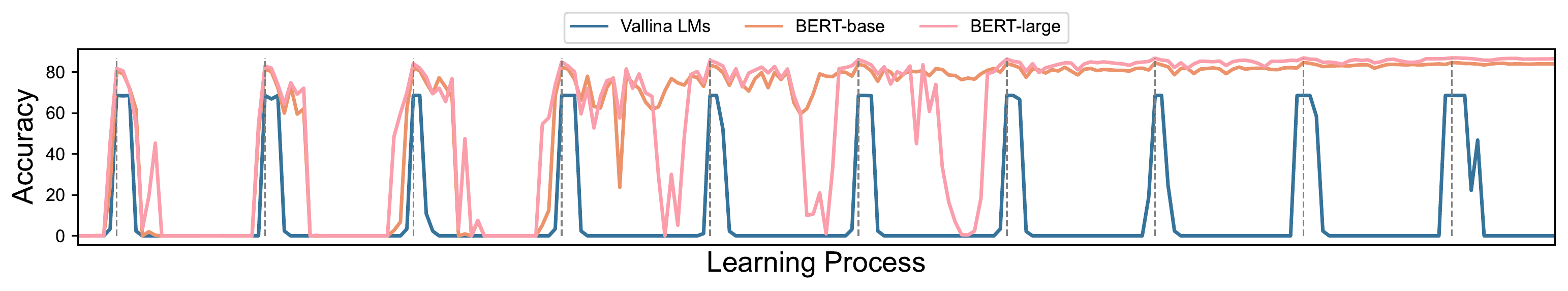}} \\
\end{figure*}

\begin{figure*}[!tp]
    \centering
    \subfloat[The forgetting curve of \texttt{original language of film}.]{\includegraphics[width=0.95\textwidth]{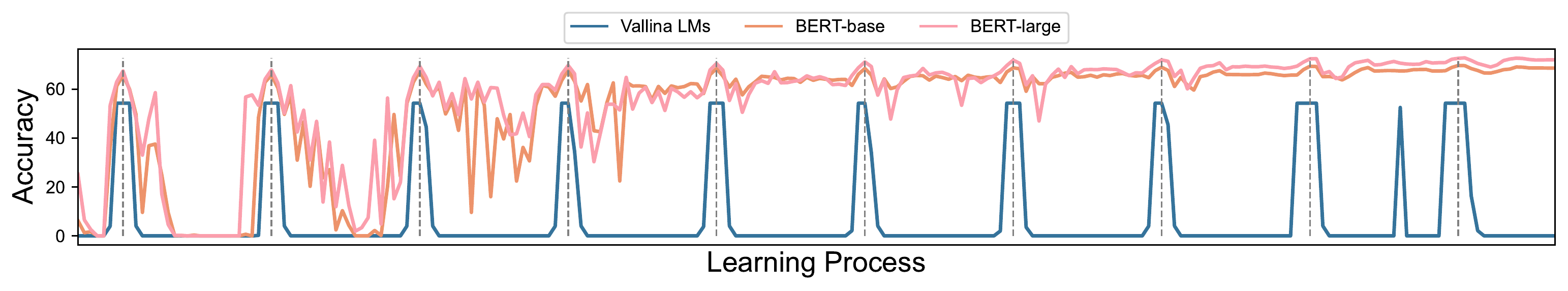}} \\
    \subfloat[The forgetting curve of \texttt{languages spoken}.]{\includegraphics[width=0.95\textwidth]{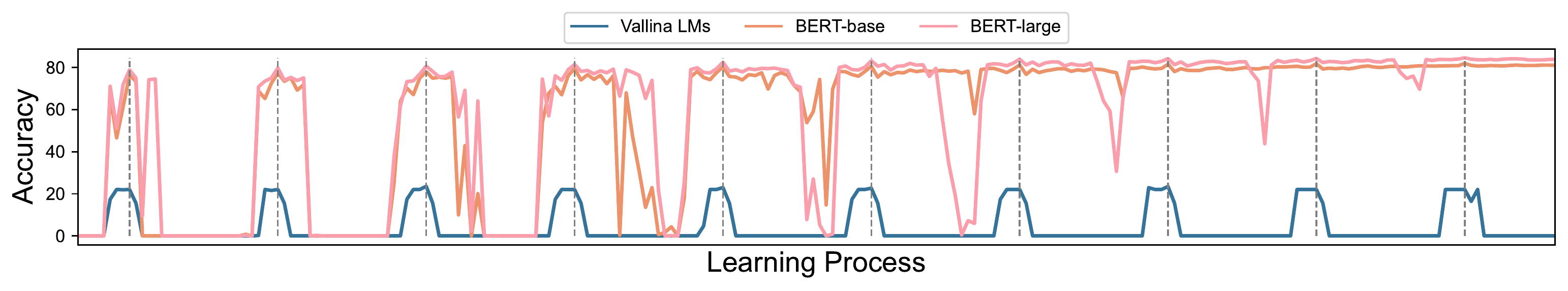}} \\
    \subfloat[The forgetting curve of \texttt{native language}.]{\includegraphics[width=0.95\textwidth]{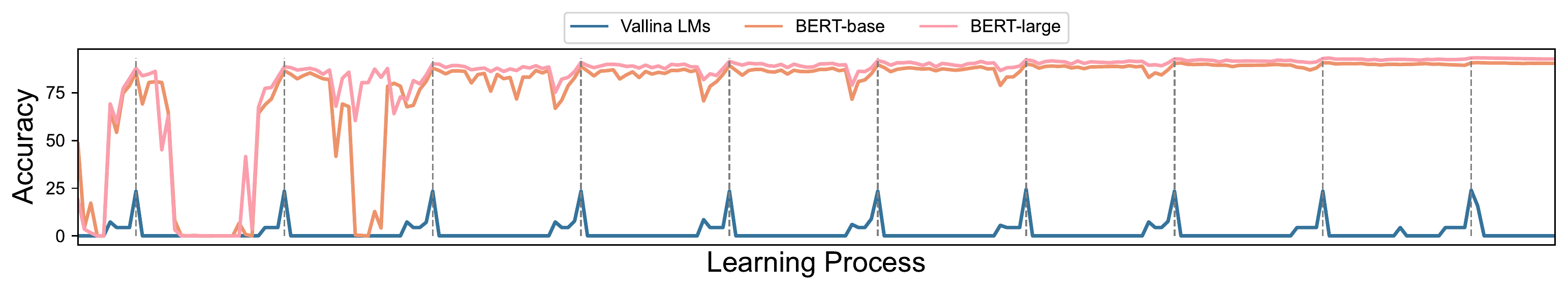}} \\
    \subfloat[The forgetting curve of \texttt{original broadcaster}.]{\includegraphics[width=0.95\textwidth]{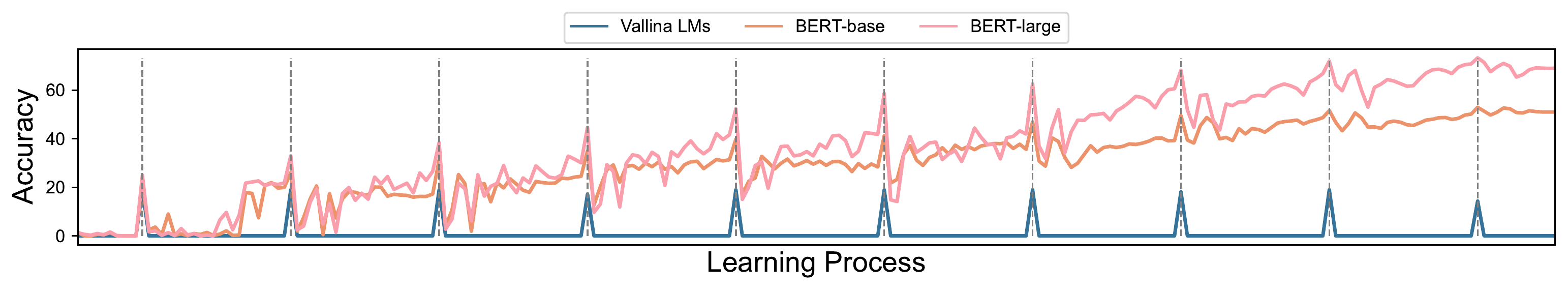}} \\
    \subfloat[The forgetting curve of \texttt{place of death}.]{\includegraphics[width=0.95\textwidth]{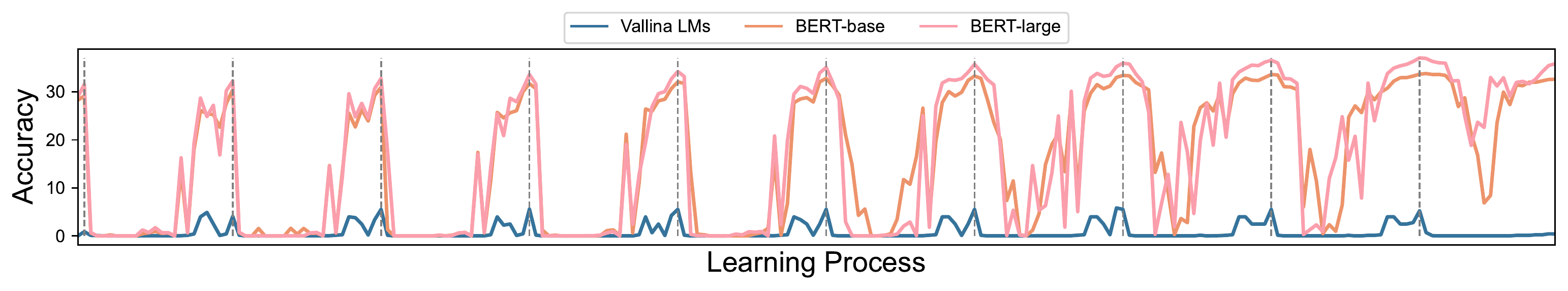}} \\
    \subfloat[The forgetting curve of \texttt{developer}.]{\includegraphics[width=0.95\textwidth]{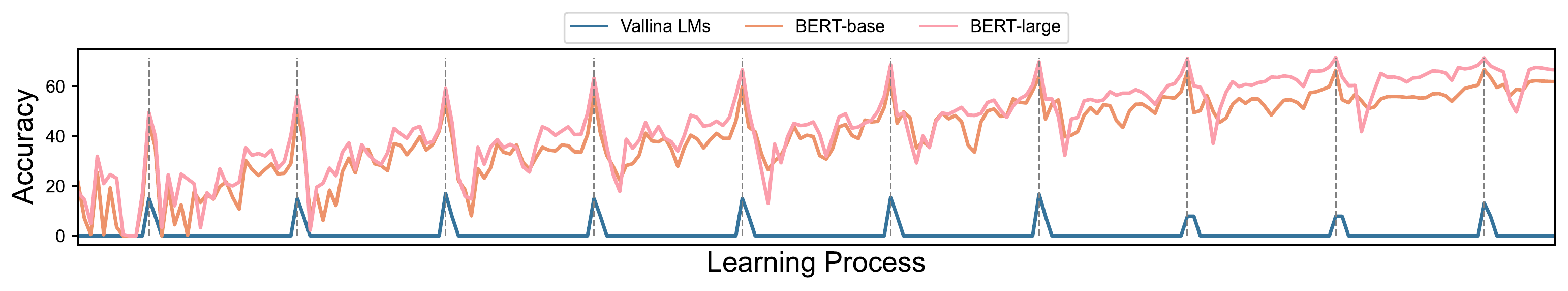}} \\
\end{figure*}

\begin{figure*}[!tp]
    \centering
    \subfloat[The forgetting curve of \texttt{manufacturer}.]{\includegraphics[width=0.95\textwidth]{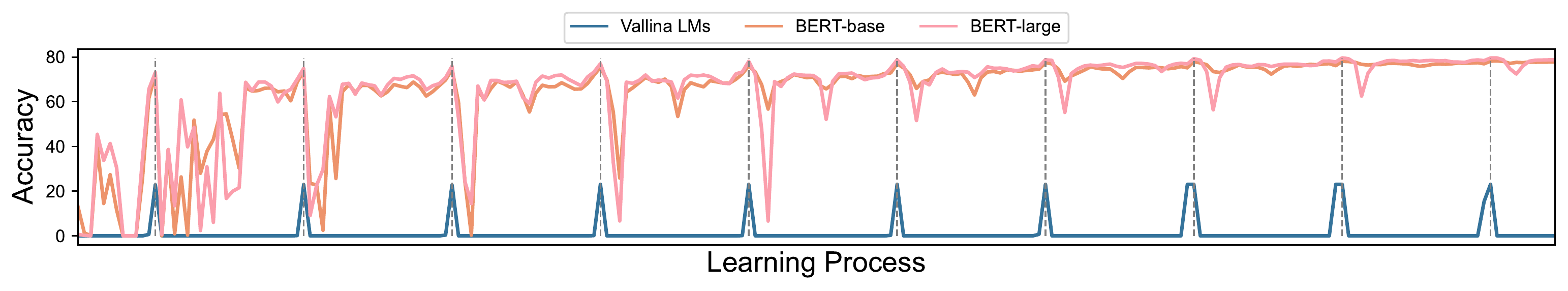}} \\
    \subfloat[The forgetting curve of \texttt{country of origin}.]{\includegraphics[width=0.95\textwidth]{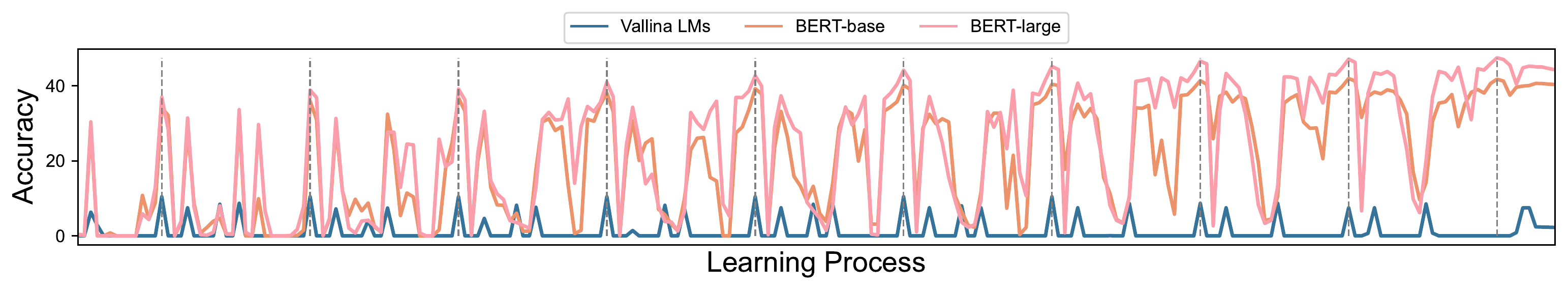}} \\
    \subfloat[The forgetting curve of \texttt{record label}.]{\includegraphics[width=0.95\textwidth]{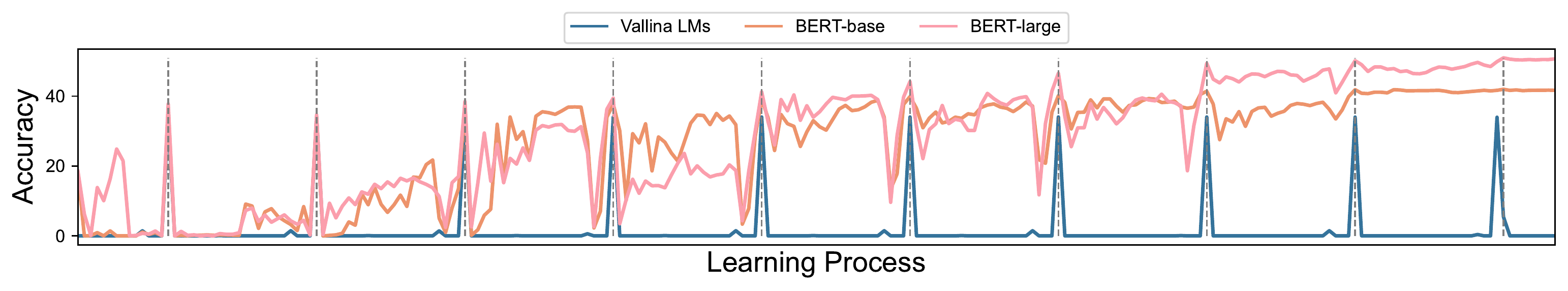}} \\
    \subfloat[The forgetting curve of \texttt{continent}.]{\includegraphics[width=0.95\textwidth]{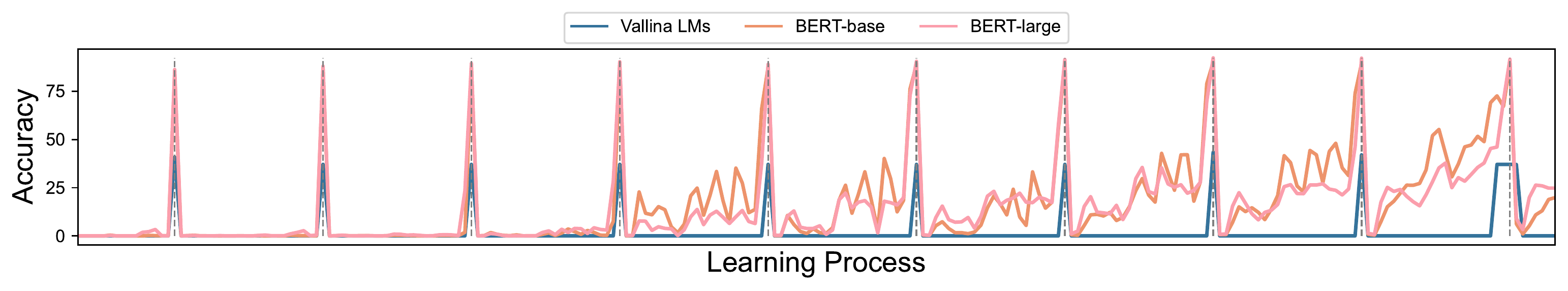}} \\
    \subfloat[The forgetting curve of \texttt{location}.]{\includegraphics[width=0.95\textwidth]{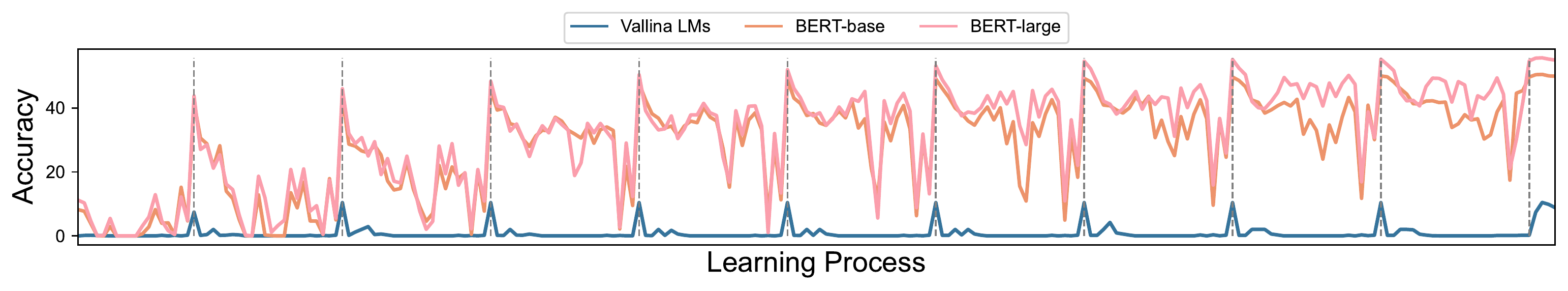}} \\
    \subfloat[The forgetting curve of \texttt{headquarters location}.]{\includegraphics[width=0.95\textwidth]{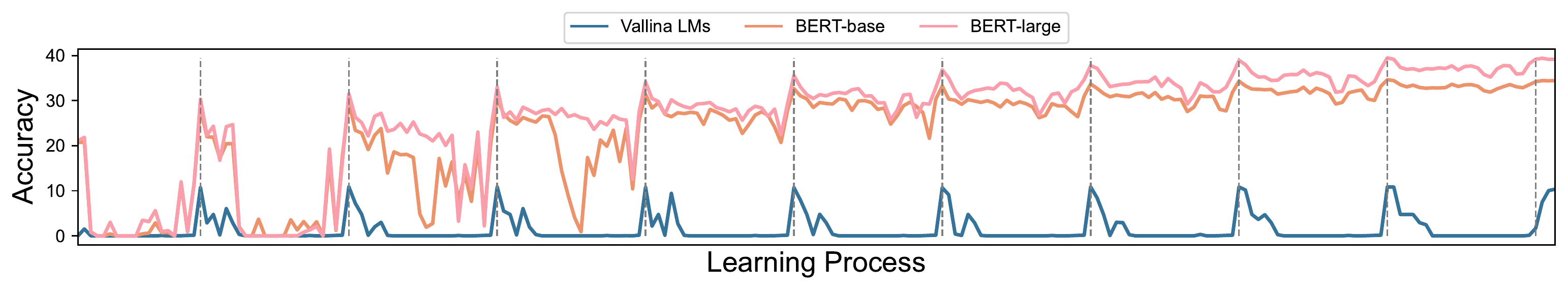}} \\
\end{figure*}

\begin{figure*}[!tp]
    \addtocounter{figure}{3}
    \centering
    \subfloat[The forgetting curve of \texttt{capital}.]{\includegraphics[width=0.95\textwidth]{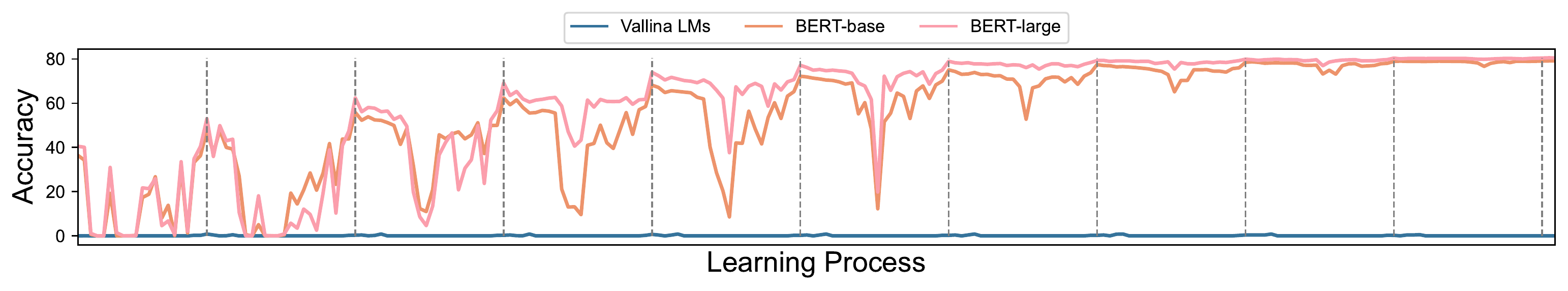}} \\
    \subfloat[The forgetting curve of \texttt{location of formation}.]{\includegraphics[width=0.95\textwidth]{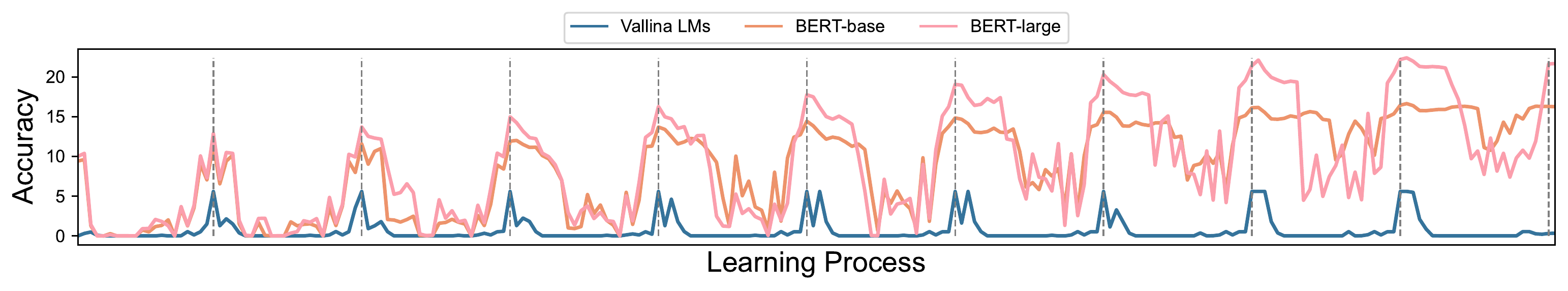}} \\
    \subfloat[The forgetting curve of \texttt{capital of}.]{\includegraphics[width=0.95\textwidth]{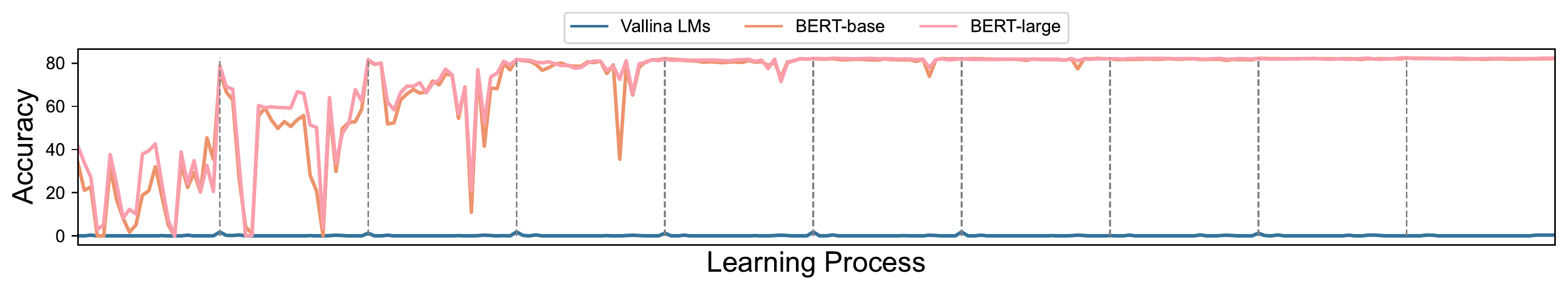}} \\
    \caption{The forgetting curves of other knowledge types on both vanilla language model, pre-trained BERT-base and pre-trained BERT-large language models.}
    \label{fig:more_forgetcurve}
\end{figure*}

\begin{figure*}[!tp]
    \centering
    \subfloat[The forgetting curve of \texttt{capital of}.]{\includegraphics[width=0.95\textwidth]{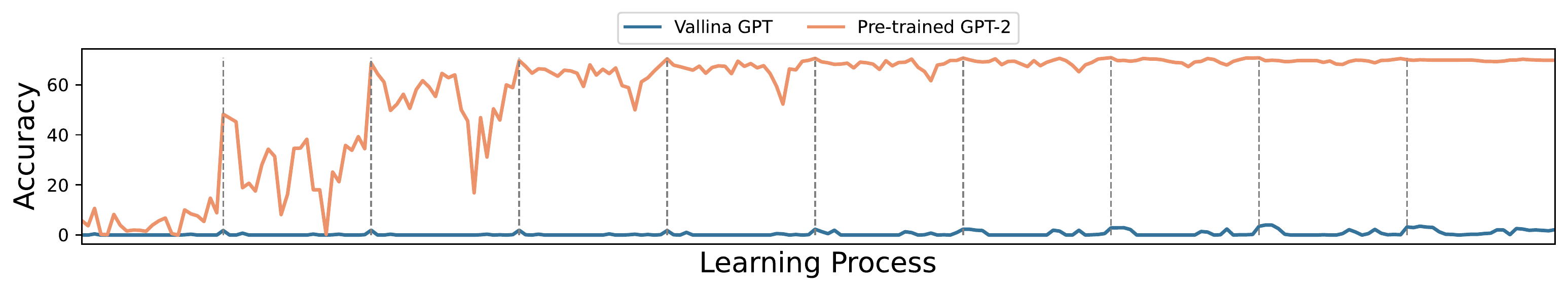}} \\
    \subfloat[The forgetting curve of \texttt{developer}.]{\includegraphics[width=0.95\textwidth]{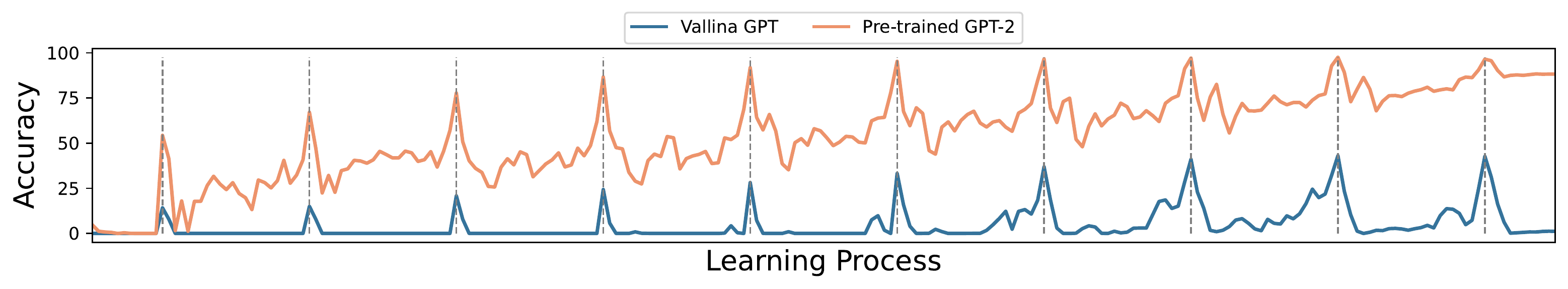}} \\
    \subfloat[The forgetting curve of \texttt{original language}.]{\includegraphics[width=0.95\textwidth]{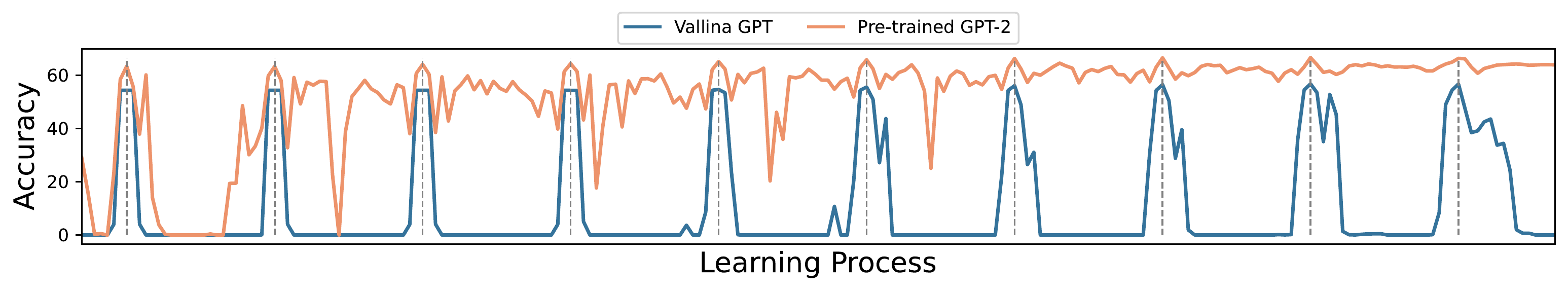}} \\
\end{figure*}

\begin{figure*}[!tp]
    \ContinuedFloat
    \addtocounter{figure}{1}
    \centering
    \subfloat[The forgetting curve of \texttt{capital}.]{\includegraphics[width=0.95\textwidth]{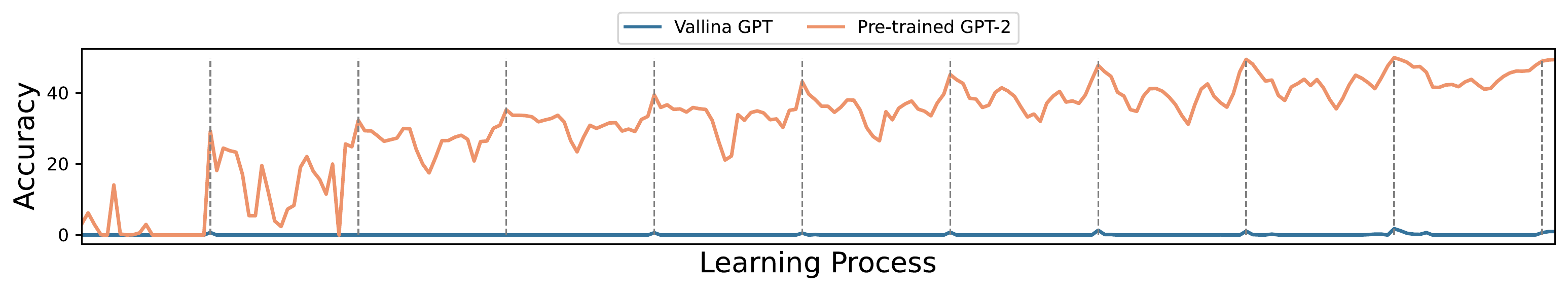}} \\
    \subfloat[The forgetting curve of \texttt{language of work or name}.]{\includegraphics[width=0.95\textwidth]{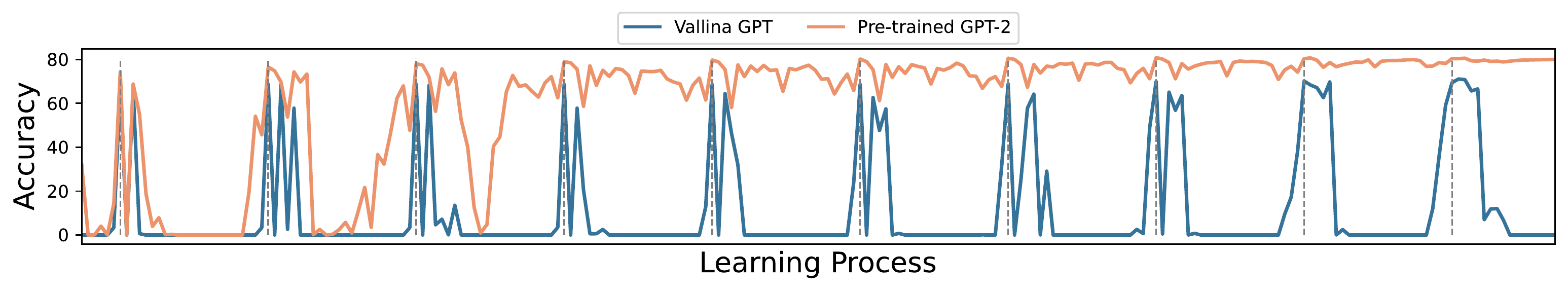}} \\
    \subfloat[The forgetting curve of \texttt{place of death}.]{\includegraphics[width=0.95\textwidth]{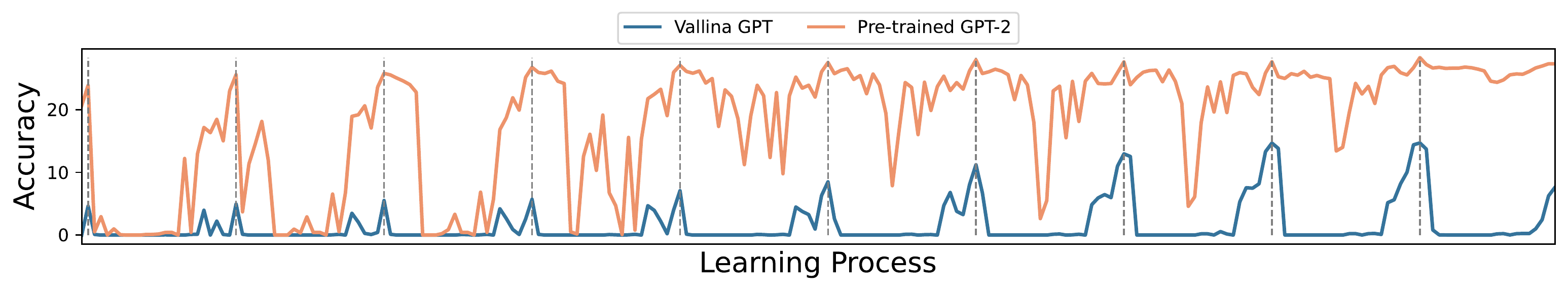}} \\
    \subfloat[The forgetting curve of \texttt{official language}.]{\includegraphics[width=0.95\textwidth]{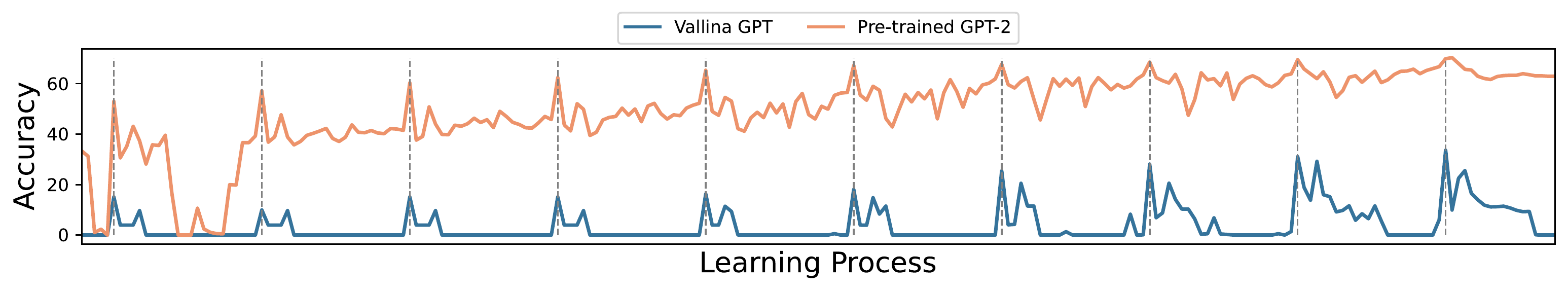}} \\
    \subfloat[The forgetting curve of \texttt{field of work}.]
    {\includegraphics[width=0.95\textwidth]{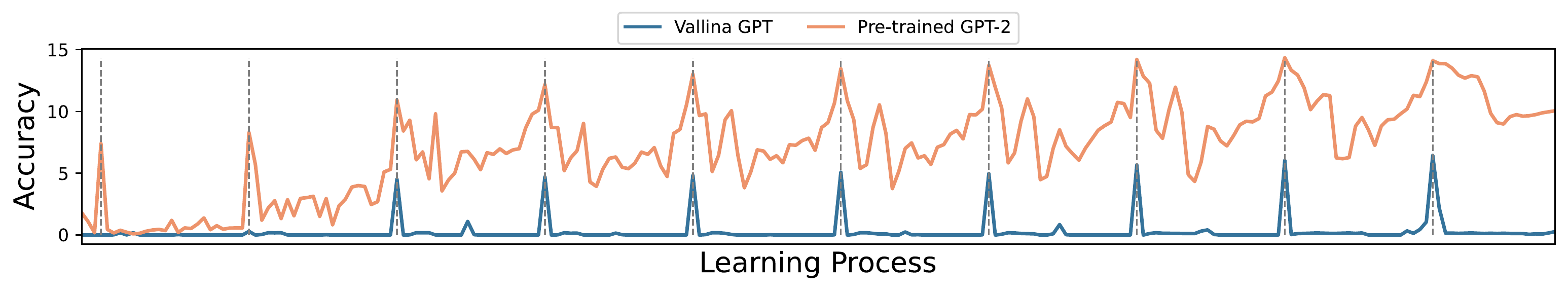}} \\
    \subfloat[The forgetting curve of \texttt{languages spoken, written or signed}.]{\includegraphics[width=0.95\textwidth]{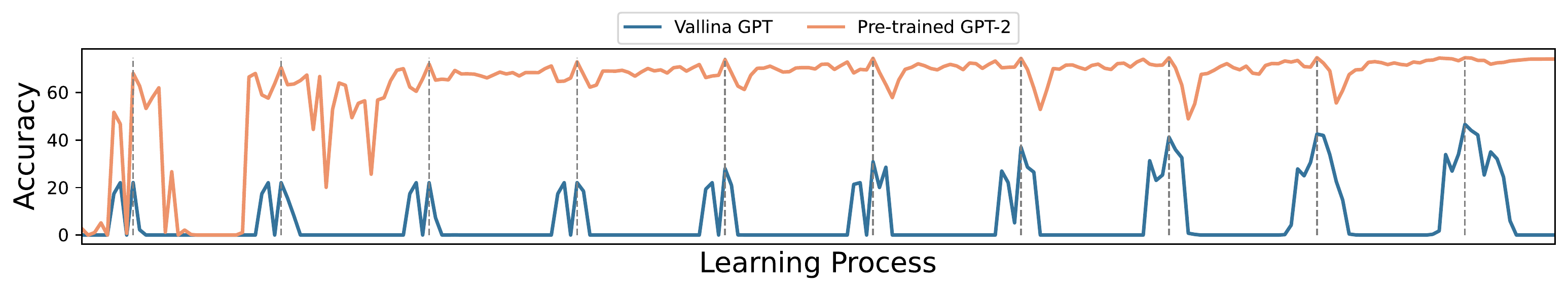}} \\
    \caption{The forgetting curves on vanilla and pre-trained GPT-2 language models.}
    \label{fig:gpt_curve}
\end{figure*}

\begin{figure*}[!tp]
    \centering
    \subfloat[\texttt{place of birth}.]{\includegraphics[width=\textwidth]{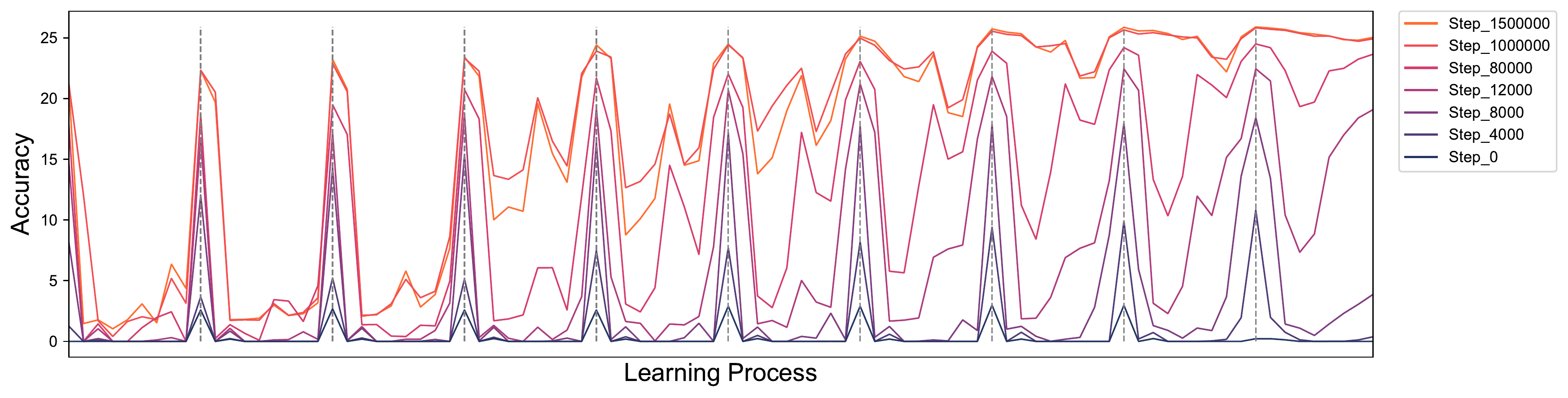}} \\
    \subfloat[\texttt{country of citizenship}.]{\includegraphics[width=\textwidth]{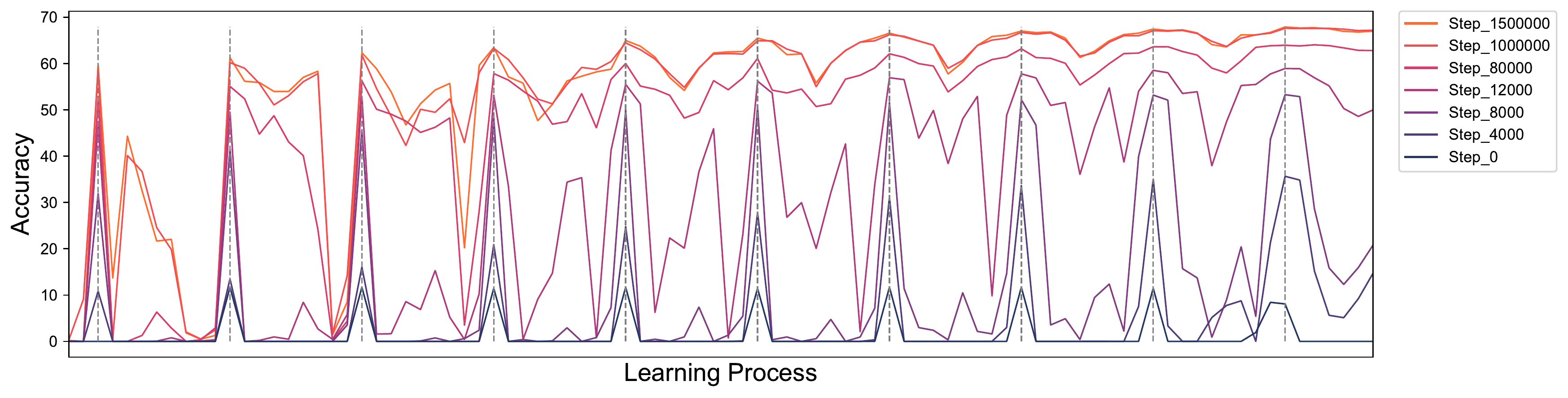}} \\
    \subfloat[\texttt{occupation}.]{\includegraphics[width=\textwidth]{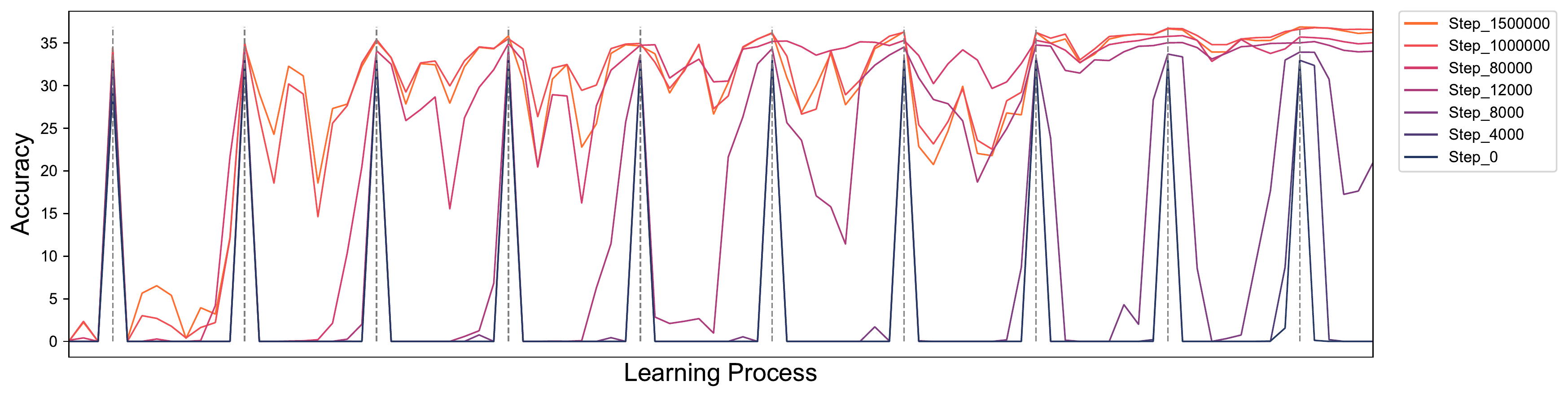}} \\
    \subfloat[\texttt{official language}.]{\includegraphics[width=\textwidth]{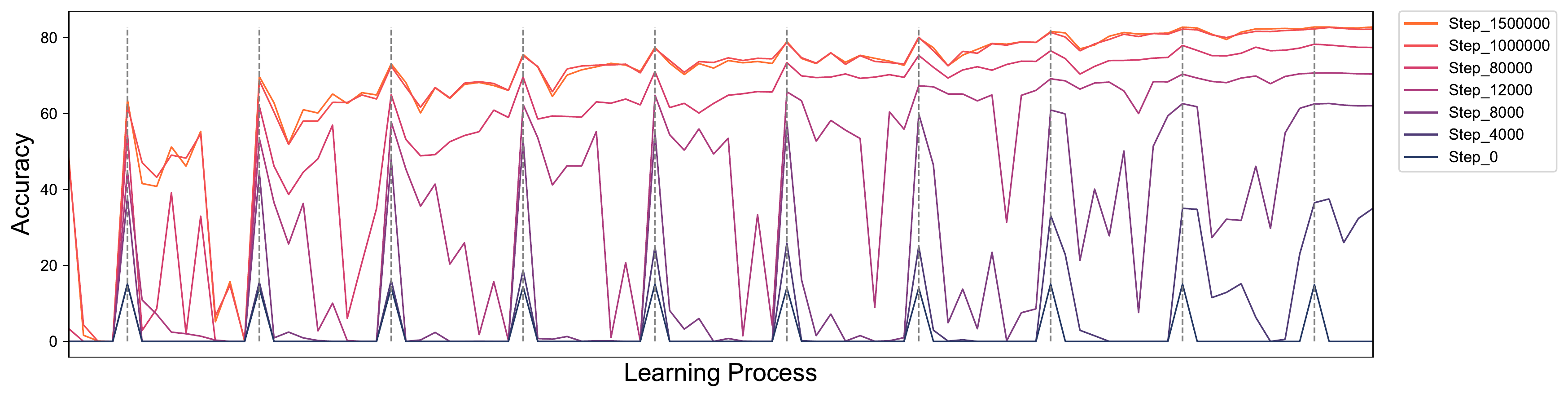}} \\
\end{figure*}

\begin{figure*}[!tp]
    \ContinuedFloat
    \addtocounter{figure}{1}
    \centering
    \subfloat[\texttt{original broadcaster}.]{\includegraphics[width=\textwidth]{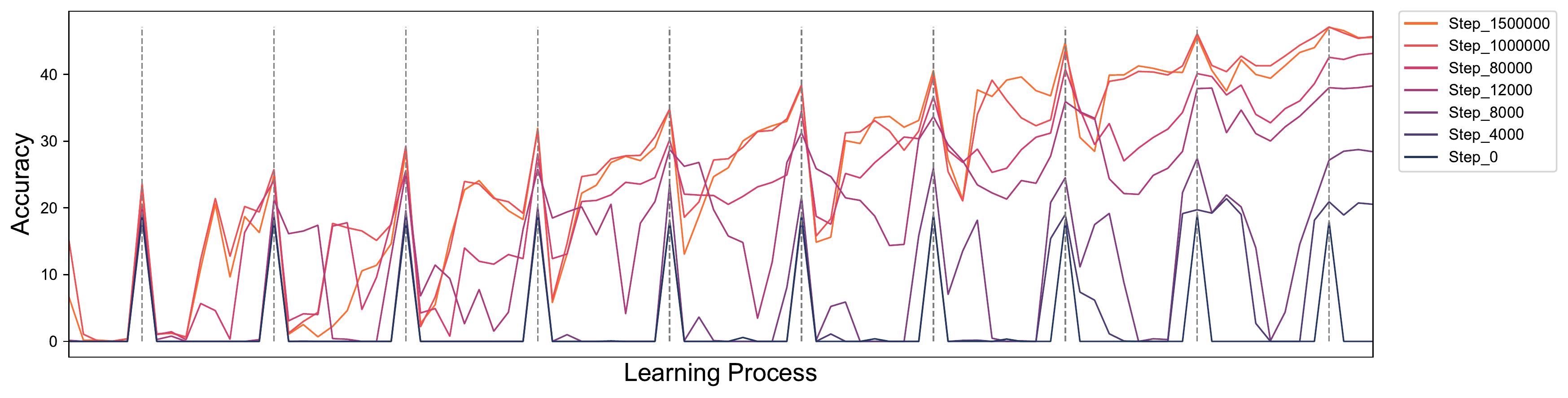}} \\
    \subfloat[\texttt{developer}.]{\includegraphics[width=\textwidth]{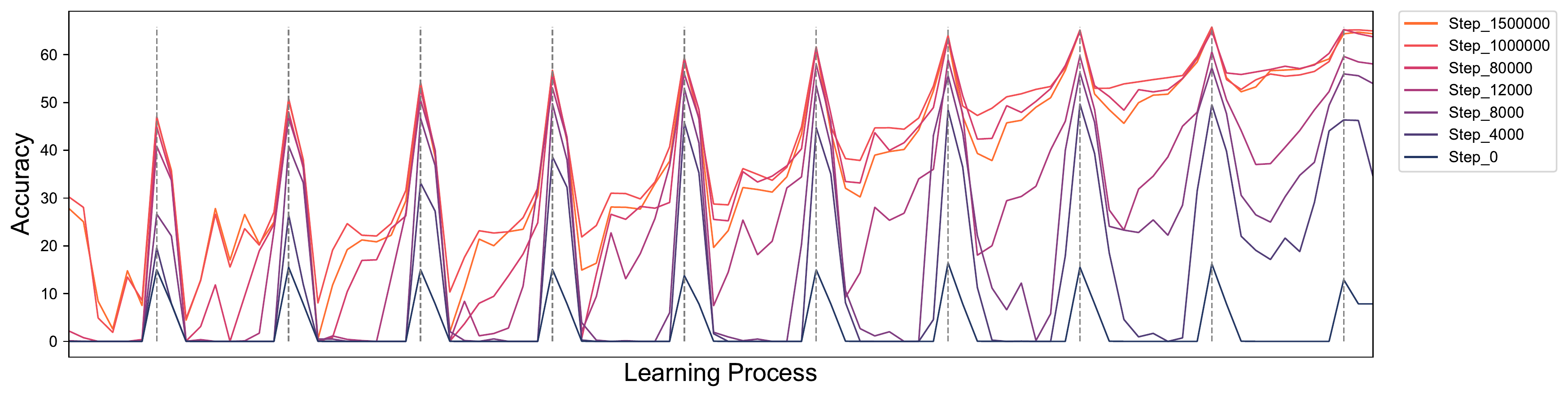}} \\
    \subfloat[\texttt{manufacturer}.]{\includegraphics[width=\textwidth]{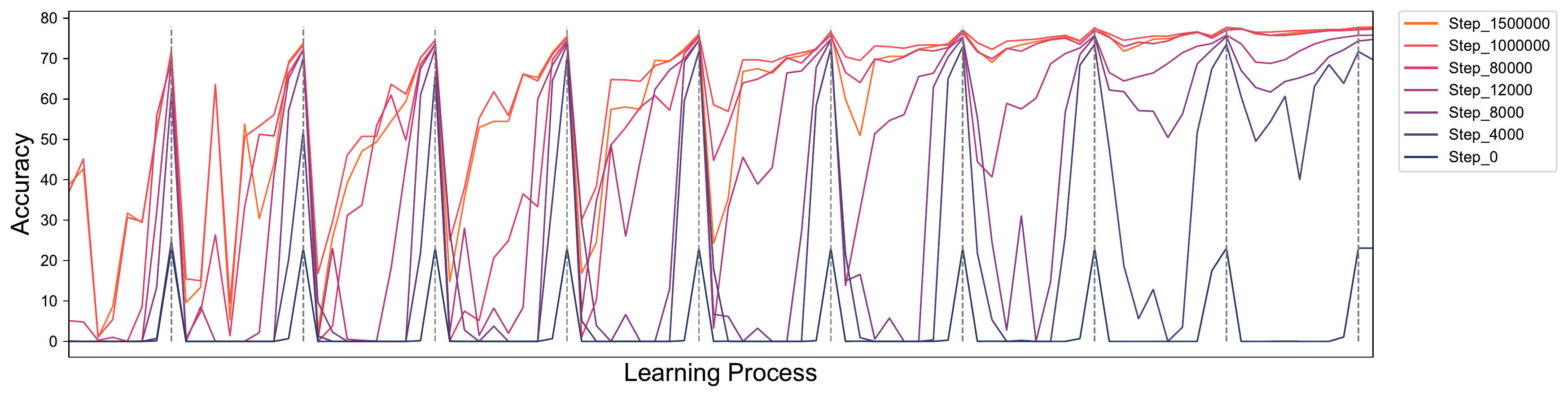}} \\
    \subfloat[\texttt{record label}.]{\includegraphics[width=\textwidth]{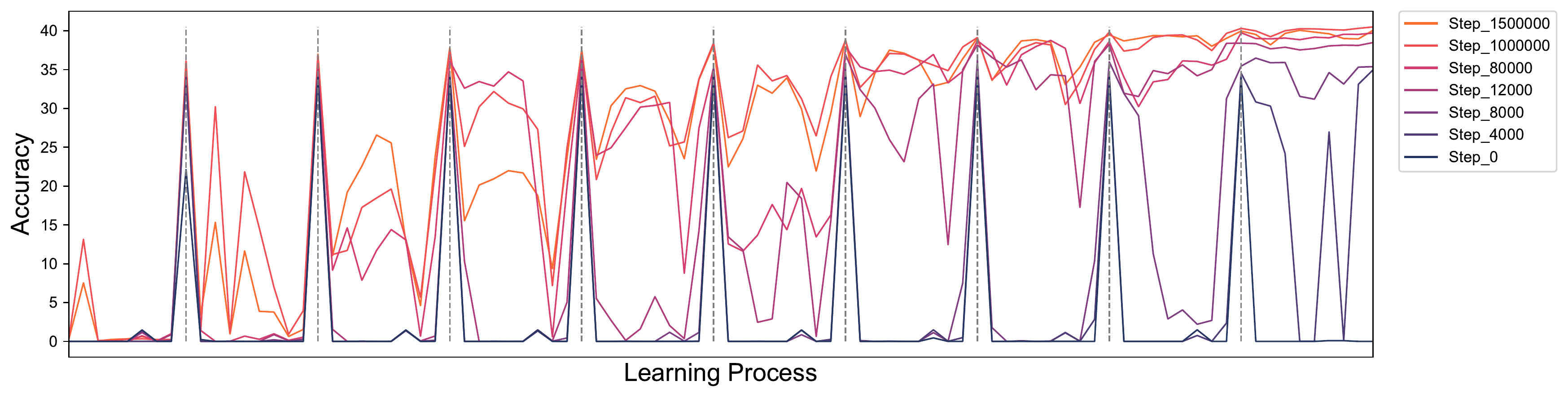}} \\
    \caption{The forgetting curves of more knowledge types on several pre-training language model checkpoints with different pre-training steps, and demonstrate the same conclusion in Section~\ref{sec:pretrain}.}
    \label{fig:more_pretraining_curve}
\end{figure*}


\end{document}